\documentclass[a4page,twocolumn]{article}

\usepackage{geometry}
\usepackage{graphicx}
\usepackage[dvipsnames]{xcolor}
\usepackage{amssymb}
\PassOptionsToPackage{hyphens}{url}\usepackage{hyperref}
\usepackage{multirow}
\usepackage{xspace}
\usepackage{booktabs}
\usepackage{soul}
\usepackage{colortbl}

\newcommand{\eg}{\emph{e.g.}\@\xspace}
\newcommand{\ie}{\emph{i.e.}\@\xspace}
\newcommand{\etal}{\emph{et al}\@\xspace}

\newcommand{\OURD}{DISC21\@\xspace}

\newcommand{\matthijs}[1]{{\color{blue}[\textbf{Matthijs}:#1]}}

\title{The 2021 Image Similarity Dataset and Challenge}

\author{
Matthijs Douze\textsuperscript{1}
\and 
Giorgos Tolias\textsuperscript{2} 
\and 
Ed Pizzi\textsuperscript{1}
\and 
Zo\"e Papakipos\textsuperscript{1}  
\and 
Lowik Chanussot\textsuperscript{1} 
\and
Filip Radenovic\textsuperscript{1} 
\and 
Tomas Jenicek\textsuperscript{2}
\and
Maxim Maximov\textsuperscript{3}
\and 
Laura Leal-Taix\'e\textsuperscript{3}
\and 
Ismail Elezi\textsuperscript{3}
\and 
Ondřej Chum\textsuperscript{2}
\and 
Cristian Canton Ferrer\textsuperscript{1} 
\\ 
~\\
\textsuperscript{1}Facebook AI\\
\textsuperscript{2}VRG, Faculty of Electrical Engineering, Czech Technical University in Prague\\
\textsuperscript{3}Technical University Munich\\
}
\date{}

\begin{document}

\maketitle

\begin{abstract}
This paper introduces a new benchmark for large-scale image similarity detection. 
This benchmark is used for the Image Similarity Challenge at NeurIPS'21 (ISC2021).
The  goal is to determine whether a query image is a modified copy of any image in a reference corpus of size 1~million. 
The benchmark features a variety of image transformations such as automated transformations, hand-crafted image edits and machine-learning based manipulations. 
This mimics real-life cases appearing in social media, for example for integrity-related problems dealing with misinformation and objectionable content.
The strength of the image manipulations, and therefore the difficulty of the benchmark, is calibrated according to the performance of a set of baseline approaches. 
Both the query and reference set contain a majority of ``distractor'' images that do not match, which corresponds to a real-life needle-in-haystack setting, and the evaluation metric reflects that.
We expect the \OURD benchmark to promote image copy detection as an important and challenging computer vision task and refresh the state of the art. 
Code and data are available at \url{https://github.com/facebookresearch/isc2021}.
\end{abstract}

\vspace{-12pt}
\section{Introduction}

Assessing whether an image is an edited copy of another source image or whether two images are edited copies of the same source image is a central task in the context of preserving information legitimacy and integrity, especially in social media~\cite{Integrity2020}.

The context is a system that tracks an image to determine its source, across platforms, through edits and re-encodings. 
There are many systems-level components of such a tracking system.
The fundamental computer vision task is to \emph{determine whether a part of an image has been copied from another image}. 

The differences between the two images are due to various kinds of image degradation, encoding artifacts, image editing and manipulations. 
Thus, it is a 2D only transformation. 
The simplest variation is image re-encoding or resizing, generating near-exact copies, while more complex changes include cropping, color variations and collages with other images.

This topic has gathered some attention over decades and it has been deemed as a mature or even solved problem. 
However, most of the state-of-the-art solutions tend to deliver unsatisfactory results with dealing with large-scale corpora.
Real-life scenarios of user-generated content involve billions to trillions of images. 
With this idea in mind, this paper proposes the image similarity dataset and competition where we present a sufficiently large and difficult dataset  to extrapolate algorithms to this operation scale. 
We call the dataset \emph{Dataset for ISC 2021}, or \OURD. 
The dataset has been constructed with practical and commonly used types of manipulations between the query image and the target ones involving geometric, hand-made and even deepfake alterations. 
In this way, \OURD provides a proxy measurement of algorithms in a real operation scenario (i.e. provenance) used in tasks like misinformation treatment, objectionable content detection and many others.

Publications on the subject of image and video copy detection are rare and not well known. 
This is because (1) the task is considered easy by the computer vision community and (2) the task is often adversarial, so organizations that use copy detection want to keep the techniques as obscure as possible. 
From a research point of view, it is useful to raise awareness in community that a ``solved problem'' like copy detection is not really solved at scale and in an adversarial setting. 
Also, there are subtle image matching problems that are not studied widely in the community and that are research problems in their own right. 
For example: determine what area of an image has been copied~\cite{tralic2013comofod,christlein2012evaluation}, or what transformation was applied to an image~\cite{wang2019detecting}.

On the other end of the semantic spectrum, it is striking that most recent unsupervised image embedding extractors (SimCLR~\cite{chen2020simple}, Momentum contrast~\cite{he2020momentum}, Deepcluster~\cite{caron2018deep}) are trained exactly to do copy detection.
Indeed, they use invariance to data augmentation as their main training criterion.
Therefore they are directly optimizing for image copy detection.
Self-supervised learning methods have not been widely applied to the task of image copy detection and we hope this dataset and competition will be an occasion to explore this direction.

The paper is organized as follows. 
Section~\ref{sec:motivation} describes the industrial context of the task. 
Section~\ref{sec:related} summarizes works on copy detection and related tasks. 
In Section~\ref{sec:dataset} the process to build the dataset is introduced, together with the evaluation metrics. 
Section~\ref{sec:baselines} reports baseline results on the dataset. 
Section~\ref{sec:rules} summarizes the rules of the ISC2021 competition.

\section{Motivation}
\label{sec:motivation}

Copy detection is widely used by Internet services, both to implement novel
product features such as reverse image search, as well as to prevent the spread
of content that have a negative social impact.

\paragraph{Use cases.}

Platforms use copy detection to apply moderator judgements to new copies of violating content, allowing them to moderate content more quickly, and at a larger scale than would be possible using manual content moderation.
Image fingerprints, such as PhotoDNA from Microsoft, are used
throughout the industry to identify images that depict child exploitation and
abuse~\cite{photodna}.
Similar techniques can be applied to identify other types of violent or self-harm imagery.

Copy detection enables rapid response to viral or emergent offensive content,
such as taking down copies of a live stream of the
Christchurch shooting~\cite{fbchristchurch}.

This allows taking action more quickly than classification models can
be updated and deployed.
Copy detection is also central to problem domains that require human judgement to identify.
Facebook uses copy detection to identify misinformation by matching
images to existing images that are associated with fact checking references~\cite{FBAI2020covidmisinfo}.

In parallel, copy detection is used to find unauthorized copies of copyrighted media.
For this task, copyrighted media to detect are identified by copyright holders.

On the negative side, one has to be aware that this technology could be used to implement extensive media censorship in some countries.

\paragraph{A solved problem?}

Image copy detection has received some attention over the years and may even be considered a solved problem.
However, two new aspects of the problem make the problem especially hard to solve: scale and new attacks.

\paragraph{Scale.}

Copy detection algorithms are often deployed at large scale, both in terms
of query volume and database size.
For example, billions of images are uploaded to Facebook systems per day.
The efficiency of matching algorithms is critical
to making large scale copy detection practical. In particular, it is important to
retrieve matches within a limited candidate volume.

In applied copy detection systems, the majority of query images
do not have matches in the index.
Thus, overall efficiency of the overall system is primarily determined by its efficiency
processing non-match queries (\eg the number of candidates retrieved).

The consequence is that copy detection typically operates in the high precision (and low recall) regime.
Since the content to be detected is low-prevalence (needle in haystack), when moving from a small-scale experimental setting like this competition to a production setting, the weight of the false positives count increases much faster than the number of false negatives.
Because of the flood of false positives due to scale, the only practical setting
is to action only on high-confidence matches.

Retroactive matching, which aims to find existing copies of newly identified violating images,
poses additional scale challenges. Retroactive matching databases can contain hundreds of billions
to trillions of image fingerprints.

\paragraph{Difficult transforms.}

Changes in user behavior have made copy detection more challenging.
Popular social media apps like Instagram and TikTok offer rich
options to edit images and videos, expanding the set of transformations accessible to typical users.
As users have moved more online activities to mobile devices,
mobile screenshots and screen captures have become common ways to share media.
Mobile %
screenshots often capture the original content in a small region of the image, capturing additional unrelated content (such as interface elements or comment text).
These changes have extended the set of common transformations that copy detection systems must be robust to.

Moreover, the setting of the copy detection task is inherently adversarial.
A user can upload content and gets immediate feedback from the platform about whether the content was blocked.
Then the user can modify it slightly and retry the upload: this is a copy detection attack.

\paragraph{In a production system.}
We briefly describe a copy detection system deployed at Facebook, as an
example of large scale applied copy detection.
As images are uploaded to Facebook products, multiple copy detection fingerprints
are extracted, including conventional image hashes and deep learned global
embedding models~\cite{FBAI2020misinfo}. We search over a database of images
associated with moderator actions, using a distributed approximate nearest
neighbor search system based on FAISS~\cite{johnson2017billion}.
This system searches for billions of images each day against a database of
hundreds of millions of images.

\section{Related work}
\label{sec:related}

In this section we review prior work that is related to the \OURD and the related existing datasets along with the corresponding research competitions, when there are any.
\begin{figure}[t]
\begin{center}
\includegraphics[width=0.7\linewidth]{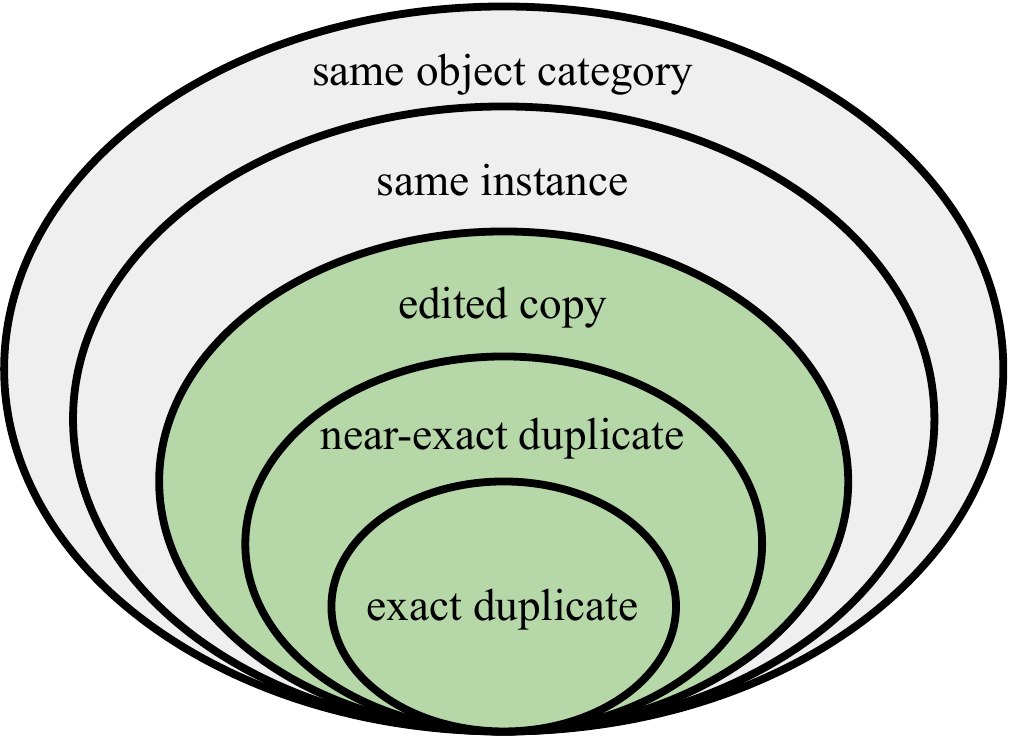}
\end{center} 
\caption{Image-pair similarity defined at different granularity levels in a nested way with the innermost level being a more restrictive one than the level above. The scope of the \OURD dataset and competition correspond to the green area.
\label{fig:levels}}
\end{figure}

\subsection{Tasks and methods}
\begin{figure}[b!]
\newcommand{\imcreds}[1]{\raisebox{0.625\depth}{\scalebox{0.4}{\rotatebox{270}{#1}}}}
\newcommand{\igp}[1]{\raisebox{-1cm}{\framebox{\includegraphics[width=0.99\linewidth,trim=50 300 100 200,clip]{figs/pres_p#1.pdf}}}}
\centering
same category\\
\igp{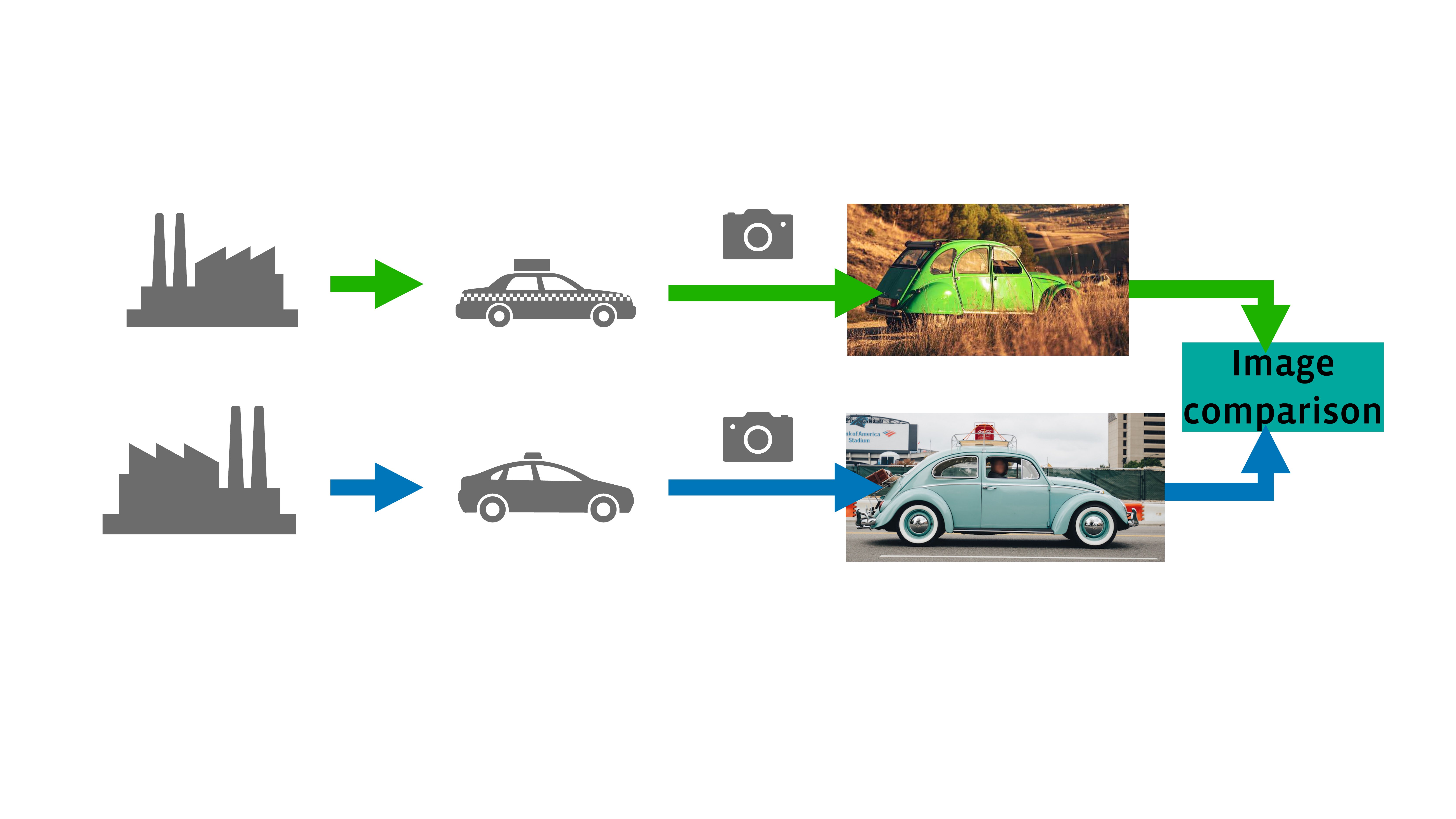}\imcreds{DaYsO on Unsplash, Wes Hicks on Unsplash}
 \\\vspace{5pt}
same instance \\
 \igp{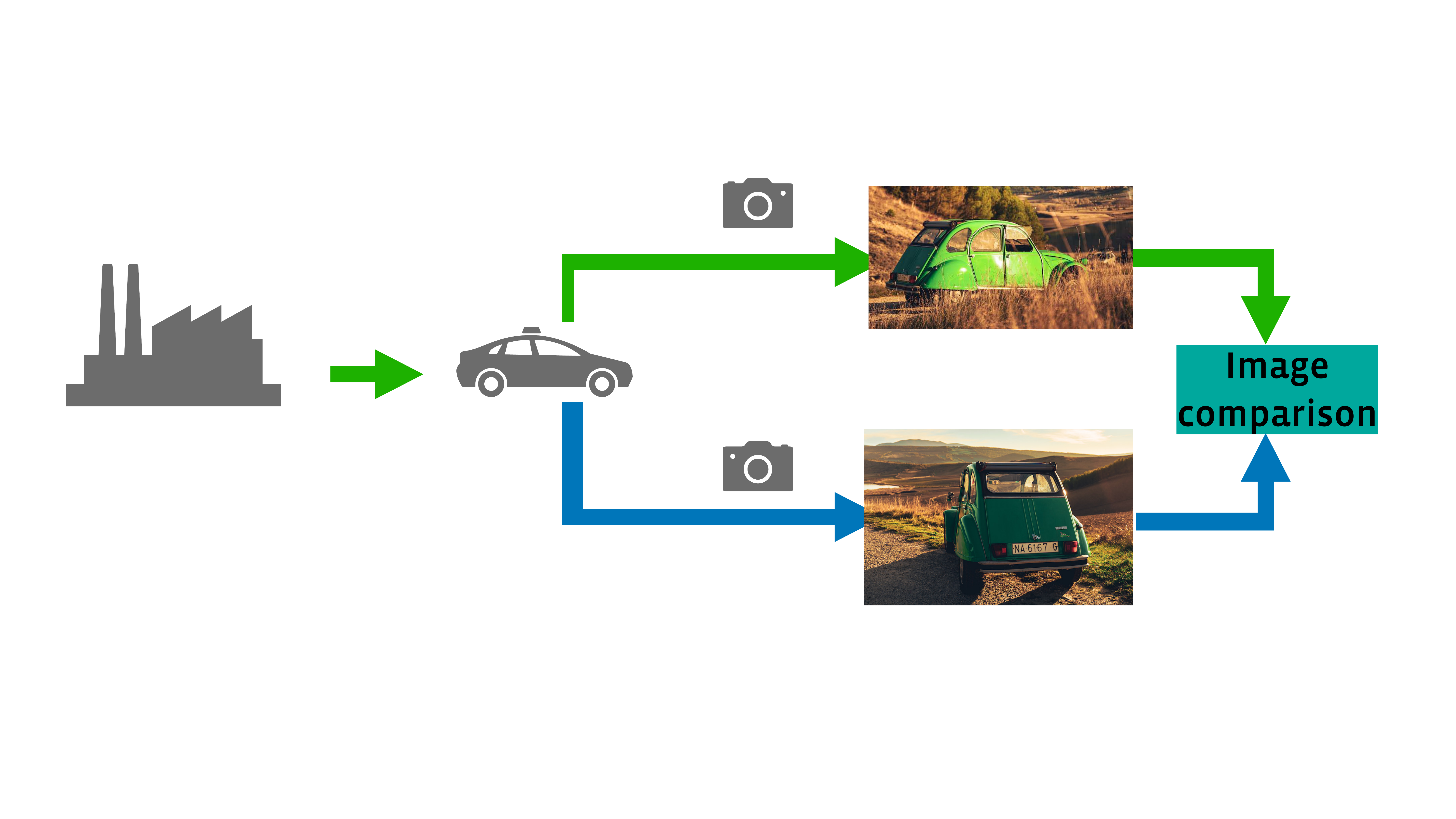} \\\vspace{5pt}
edited copy \\
 \igp{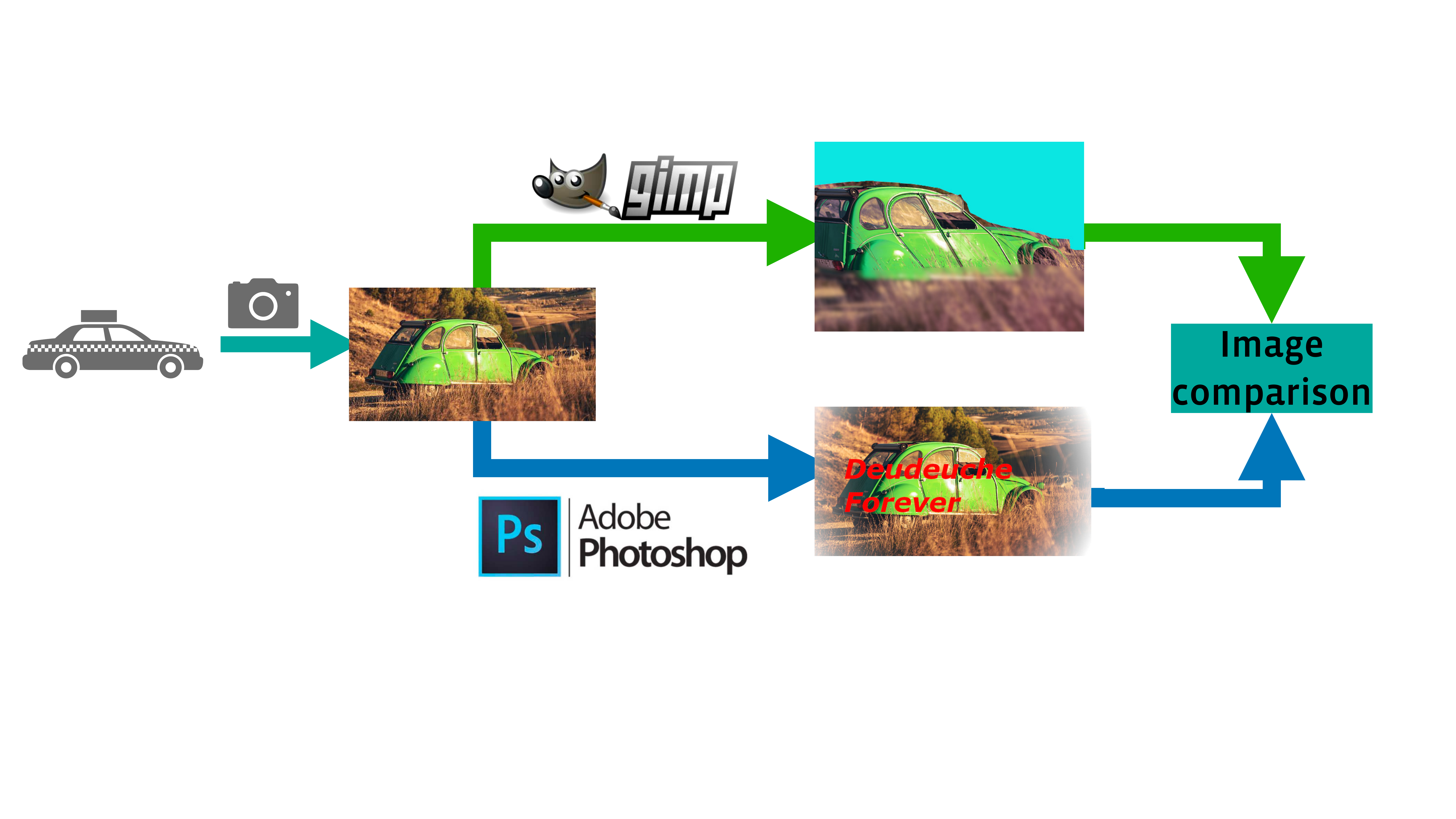} \\
\caption{\label{fig:levels_example}
Real-world imaging process that generates image pairs that are considered similar at different granularity levels.
}
\end{figure}

Two images can be considered, \ie labeled as, similar with criteria defined at different granularity levels as these are shown in Figure~\ref{fig:levels} with examples in Figure~\ref{fig:levels_example}. 
The most restrictive case corresponds to the exact duplicates, where images are bit-wise the same. 
Therefore, comparing images is trivial and reduces down to to comparing bit strings which is applied at a large scale via file comparison and hashing, \eg with MD5 hashes.
Near-exact duplicates correspond to changes that are nearly human imperceptible such as compression distortion. 
Edited copies, which are the main focus of \OURD, correspond to an image pair where both images are edited version of the same source image.
Then, there exist the cases of images showing the same instance, \eg a particular object, or two objects belonging to the same category, corresponding to the task of instance-level recognition (ILR) or category-level recognition (CLR), respectively.

We review approaches for Copy Detection (CD), but also copy retrieval due to overlap of approaches, in the domain of images and videos.
In retrieval, the objective is to rank all database images with respect to the similarity to a query image. There is no need for the similarity score to be comparable across different queries;  standard retrieval evaluation metrics perform on a per query basis. 
In detection though, the objective is to provide a list of detected copies, typically obtained by thresholding a confidence values that a query image is a copy of a source image. The confidence value  should be comparable across queries; an appropriate evaluation metric should perform across all queries jointly. 
We additionally review approaches for ILR, as a neighboring level in the definitions of image similarity, and discuss cases for the tasks of retrieval, classification, and detection. 
ILR approaches are often directly transferable to copy detection and retrieval, which is also the case in the set of baseline approaches discussed in Section~\ref{sec:baselines}. 
Besides, there is an overlap between CD and ILR, where the image is \eg printed out and captured by a camera: in that case there is a common source image but it is still a 3D image capture.

\begin{table*}
\small
\centering
\def\sp{\hspace{5pt}}
\def\spp{\hspace{10pt}}
\begin{tabular}{l@{\sp}c@{\sp}c@{\sp}c@{\sp}r@{\sp}r@{\spp}r@{\spp}c@{\sp}c@{\sp}c@{\sp}}
\hline
\multirow{2}{*}{Dataset}   	&   \multirow{2}{*}{Domain}    &  \multirow{2}{*}{Task} & \multirow{2}{*}{Edit type}   &  \multicolumn{3}{c}{Dataset size} & \multirow{2}{*}{Year} & \multirow{2}{*}{Comp.} \\
\cmidrule(rr){5-7}
 & & & & ~Queries~ & ~~~QueriesD & Database~ & & \\
 \hline \hline
Muscle-VCD~\cite{law2007muscle} & Video & CD & U  & 18 & - & 101 & 2007 & No \\
CCWeb~\cite{wu2007ccweb}    			& Video & CD & U  & 24 & - & 13k & 2007  & No \\
Trecvid~\cite{over2013trecvid}    & Video & CD & S    & 11k & - & 11k & 2008  & Yes \\
UQ\_Video~\cite{song2011multiple}    & Video & CD & U   & 24 & - & 167k & 2011 & No \\ 
EVE~\cite{revaud2013event} & Video  & ILR & -& 620 & - & 2.4k (+100k) & 2013 & No \\
VCDB~\cite{jiang2014vcdb}   			& Video & CD & U  & 528 & - & 100k & 2014  & No \\
FIVR~\cite{kordopatis2019fivr} & Video & ILR+CD &U & 100 & -  & 226k & 2019 & No \\ \hline
UKBench~\cite{nister2006scalable} & Image & ILR &-& 10.2k & - & 10.2k & 2006 & No \\
Oxford~\cite{philbin2007object}    & Image & ILR &-& 55 & - & 5k  (+100k) & 2007  & No \\
Paris~\cite{philbin2008lost}    & Image & ILR &-& 55 & - & 6k (+100k) & 2008  & No \\
Holidays~\cite{jegou2008hamming}    & Image & ILR &-& 500 &  & 1941 & 2008  & No \\
Copydays~\cite{Douze2009EvaluationOG}    & Image & CD &U+S & 157 & -  & 3k & 2009  & No \\
Instre~\cite{wang2015instre}    & Image & ILR &-& 1250 & - & 27k & 2015  & No \\
GLDv1~\cite{noh2017large}    & Image & ILR& -& 1.3k & 117k & 1.2M/1.1M & 2017  & Yes \\
ROxford~\cite{radenovic2018revisiting}  & Image & ILR& - & 70 & - & 5k (+1M) & 2018  & No \\
RParis~\cite{radenovic2018revisiting}  & Image & ILR&-  & 70 & - & 6k (+1M) & 2018  & No \\
GLDv2~\cite{wac+20}    & Image & ILR &-& 1.3k & 117k & 4.1M/762k & 2020  & Yes \\
\rowcolor{Apricot} \OURD    & Image & CD &U+S & 20k & 80k & 1M & 2021  & Yes \\
\hline
\end{tabular}
\caption{Datasets that are publicly available and are related to our dataset and competition. Comp.: whether there was a corresponding research competition or not. QueriesD: distractor queries.
U: user-generated transformations. S: synthesized transformations. 
\label{tab:datasets}}
\end{table*}

\paragraph{Video copy detection.}
The literature for copy detection is richer in the video domain than the image domain.
Early approaches rely on appearance-based global representation for the whole video~\cite{shen2007uqlips,wu2007ccweb,liu2007video,shen2005towards}, reducing the task to similarity estimation in high dimensional spaces.
Such approach is less effective for detecting partial copies. 
Frame-level matching is performed with the use of local descriptors only after a first filtering stage~\cite{wu2007ccweb}, or efficient local descriptor indexing and geometric matching~\cite{douze2010image}, or with compact aggregated representations~\cite{douze2010compact}.
As a post-processing step, the temporal consistency of the result by frame-level matching is checked in the form of Hough temporal transform~\cite{douze2010compact} or cast as a network flow problem~\cite{tan2009scalable}. Frame-level and temporal matching are jointly performed in the work of Poullot \etal~\cite{poullot2015temporal}.
Deep learning for video copy detection is used in recent approaches for video representation without temporal information~\cite{kordopatis2017near}, frame and temporal matching~\cite{hu2018learning} or spatio-temporal representations~\cite{kordopatis2019visil}.

\paragraph{Image copy detection:}
The road path of global descriptors~\cite{kim2003content,wan2008survey} is followed for image copy detection too, which Douze~\etal~\cite{Douze2009EvaluationOG} perform a performance evaluation.
Following the foot-steps of classical approaches in many other computer vision tasks, image copy detection is well handled by the use of local descriptors, the Bag-of-Words models and spatial constraints~\cite{zhou2016effective,zll+10}.
Deep networks are only used in straight-forward ways, \eg training in a Siamese way~\cite{Zhang2016copydetectioncnn}, which we interpret as loss of interest by the research community due to the lack of new, challenging, large benchmarks that reflect real world applications.
There is a more specialized line of work in detecting copy-move forgeries~\cite{tralic2013comofod,christlein2012evaluation}, an edit function where part of the image is copied and pasted into the same image, or estimation of the applied transformation~\cite{wang2019detecting}.

\paragraph{Instance-level recognition.}
Instance-level recognition has attracted a lot of attention and many relevant approaches exist in the literature, especially for  the task of retrieval.
These include purely global descriptors~\cite{Douze2009EvaluationOG}, local descriptor indexing and matching~\cite{Sivic2003VideoGA,jegou2008hamming}, local descriptor aggregation into global descriptors~\cite{jegou2012aggregating,pls+10}, spatial verification~\cite{philbin2007object} and query expansion~\cite{chum2007total}, among many others.
Deep learning approaches are currently dominating in this task with both global~\cite{babenko2014neural,razavian2016visual,radenovic2018fine,mmo+16} and local descriptors~\cite{noh2017large,simeoni2019local, tolias2020learning}.
Recently, more attention is paid on ILR for classification at large scale~\cite{noh2017large,tolias2020learning}, which bears similarities to the task of our work. Testing examples do not necessarily come from the same categories as the training ones, as some form of open set recognition, and classification is in practice performed~\cite{noh2017large,tolias2020learning} by verification that a query and database image pair come from the same class. 
Another example departing from instance-level retrieval is the work of Furon and J{\'e}gou~\cite{fj13}, where the focus is on detection of the relevant image pairs, which is also close to the setup of this work. 

\subsection{Datasets and  competitions}
Table~\ref{tab:datasets} summarizes existing datasets that are related to the benchmark we are setting up. 

\paragraph{Video datasets.}
Most copy detection datasets are in the video domain with some well established examples like VCDB~\cite{jiang2014vcdb} and CCWeb~\cite{wu2007ccweb}.
It is noteworthy that video copy detection is not a superset of image copy detection: 
video datasets focus on temporal degradations, such video edits, speed-up, slow-down, and, therefore, less on challenging frame-level degradations. 
In fact, due to the high information content of videos, it is easier to recognize copied videos than images~\cite{douze2010image}. 

\paragraph{Image datasets.}
Many of the prior approaches for image copy detection perform evaluation on private datasets, while the most well known publicly available dataset, Copydays~\cite{Douze2009EvaluationOG} is very small scale.
There are other datasets that focus in the domain of logos, such as Belgalogos~\cite{joly2009logo}, which is a very specialized case because of the very limited number of templates to recognize.
Dataset creation for copy detection at large scale is a tedious process, especially if user generated edited copies are included which also form the most challenging cases. 

\paragraph{ILR datasets}
exist at much larger scale, but when relying on crowd-sourced labels~\cite{wac+20} to achieve that the result might include some noise.
In this work, we introduce a dataset for image copy detection that (a) significantly exceeds the scale of previous ones, reaching the scale of ILR datasets, (b) contains a large amount of image copies including many user generated ones, and (c) includes a large number of distractor queries which is well aligned with the nature of a detection task.

\paragraph{Competitions}
There is no previous example of competition on copy detection on images. 
There was a video copy detection track in the TrecVid series of competitions  between 2008~\cite{douze2010image} and 2012~\cite{over2013trecvid}. 
The Google Landmarks Challenge is a recent example for ILR at large-scale~\cite{google2019landmarkschallenge} that is attracting a lot of attention, \ie more than hundred of actively participating teams on Kaggle.

\section{Dataset}
\label{sec:dataset}

The dataset is composed of three parts: the \emph{reference set}, also called \emph{database}, which consists of the set of the original (source) images which are then transformed into the edited copies, the \emph{query set}, which  consists of the set of transformed images, and the \emph{training set} which is of similar spirit to the reference set. 
There are two types of queries, depending on whether their source image is part of the reference set. 
For \emph{distractor queries}, the source is not in the reference set: there no copy to detect for those.

The dataset is a small-scale example of the task that is performed in a real-world industrial context.
Even though Facebook, for example, processes billions of images per day, we restrict the size of the dataset to 1~million. 
On the other hand, the transformations that are used to generate the edited copies are harder than what is typically seen in a real-world context, which makes \OURD a more interesting and attractive computer vision benchmark.
We additionally make sure that specialized approaches, such as optical character recognition (OCR) and face recognition do not significantly improve performance.

\subsection{Data sources}

The reference set  
is based on two sources: the YFCC100M~\cite{Thomee2016YFCC100MTN} dataset\footnote{This data source contains significantly less full images of text than what is seen typically on social media platforms.}, and the DeepFake Detection Challenge (DFDC) dataset~\cite{DFDC2020}. 
The images are chosen so that they can be redistributed without legal implications.
We choose images from YFCC100M which are under 
permissive creative commons license types and which do not contain any identifiable human; some images depict humans in the distance such that they are not identifiable, corresponding to an area no larger than $0.5\%$ of the whole image area.
Nevertheless, social media images contain many images of humans. Therefore, to make the dataset more realistic, we further add a smaller number of images which depict humans from a previous competition organized by Facebook, the DeepFake Detection Challenge (DFDC). The images we use are frames from DFDC videos of paid actors who gave their permission for their video recordings to be used. 
Around half of the DFDC images that we use are processed with a face swapping algorithm (``deepfake'')%
~\cite{CIAGAN} to manipulate the faces before applying image copy transformations. 
In this way, the transformed image and the original image cannot be easily matched using face verification techniques. 

\paragraph{Pre-processing.}

Face detection is performed on the DFDC source images in order to keep a close crop, with some randomness, of the faces. 
We do this because most DFDC videos, therefore the images too, are taken at a distance from the actors, while images on photo sharing sites are more often selfies.

\begin{figure}
\newcommand{\imcreds}[1]{\raisebox{\depth}{\scalebox{0.4}{\rotatebox{270}{Flickr / #1}}}}
\newcommand{\imcredsD}[1]{\raisebox{\depth}{\scalebox{0.4}{\rotatebox{270}{DFDC image #1}}}}
    \centering
    \newcommand{\igdup}[1]{\includegraphics[width=0.47\linewidth]{#1}}
    \begin{tabular}[b]{cc}
    \igdup{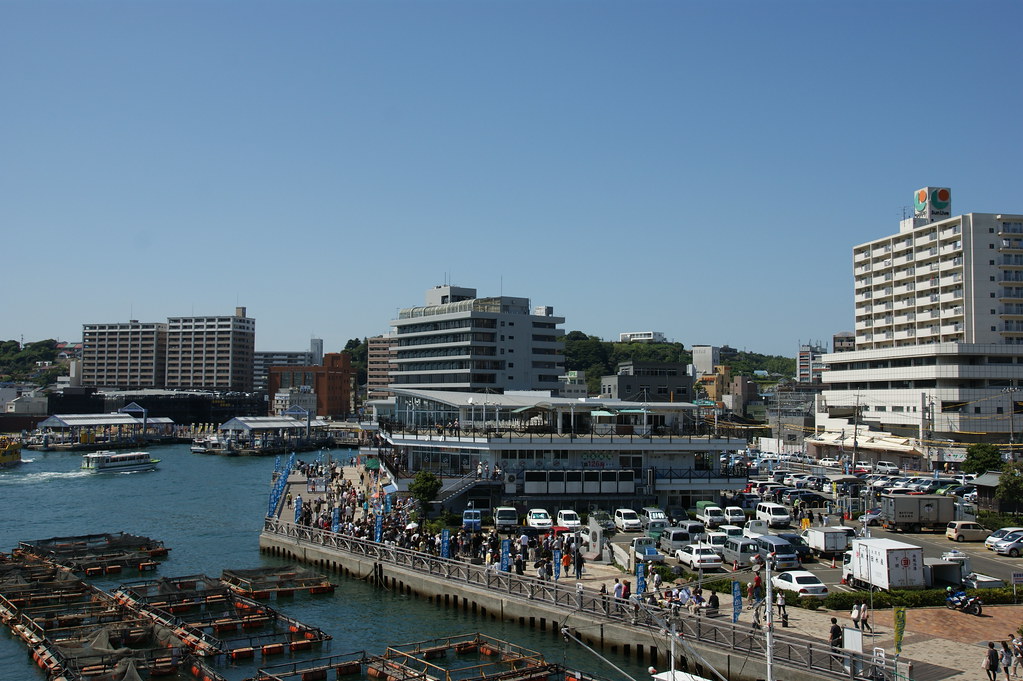} &
    \igdup{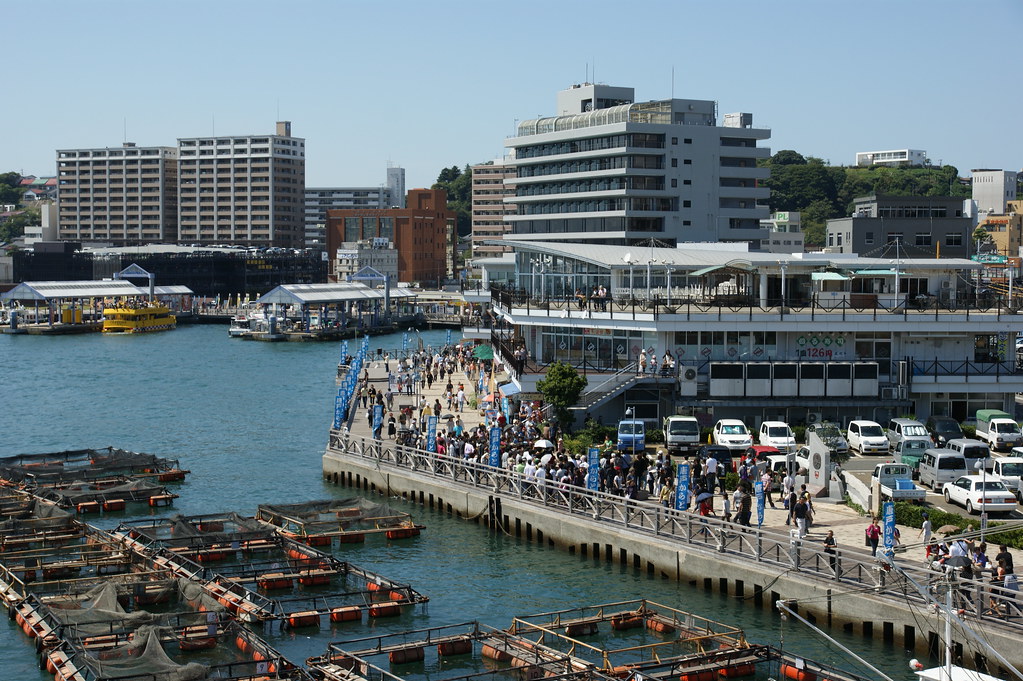}\imcreds{m-takagi}\\
    \igdup{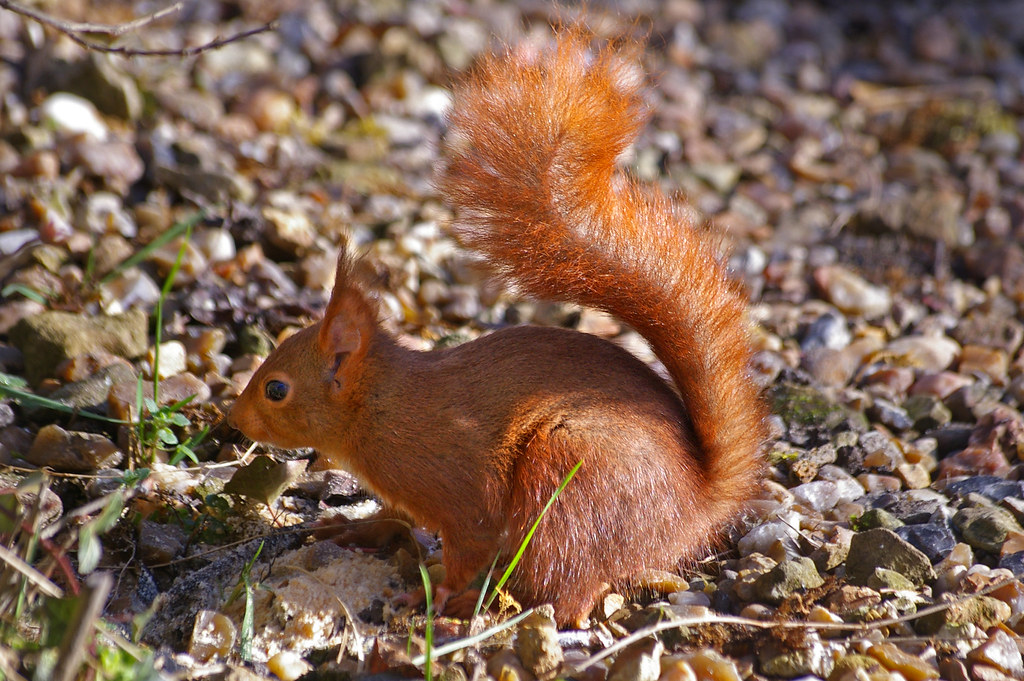}& 
    \igdup{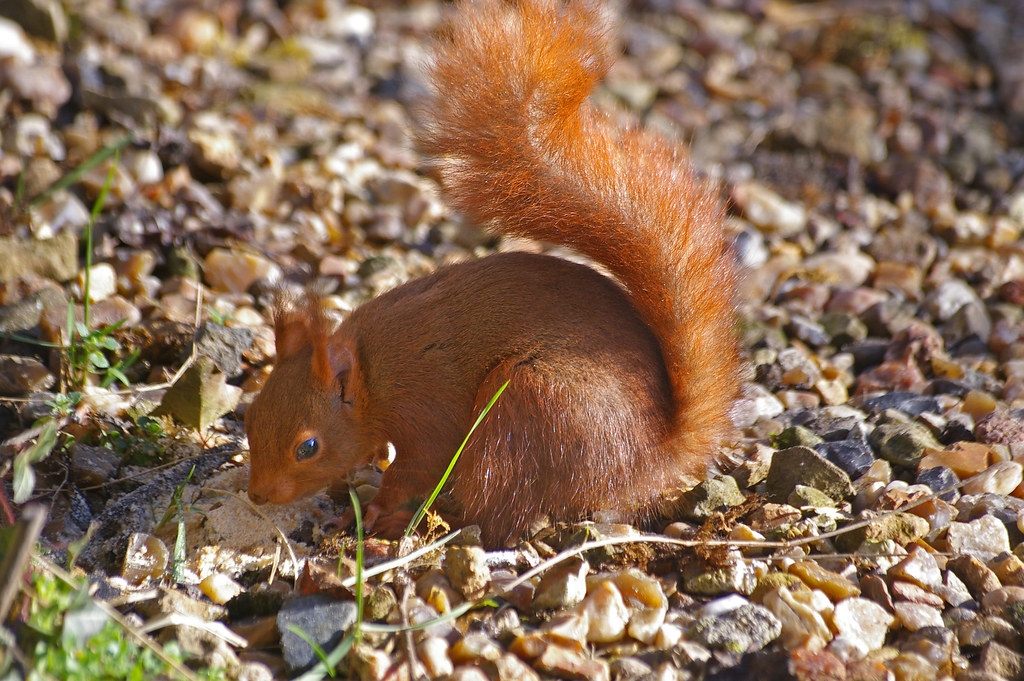}\imcreds{sybarite48}\\
    \end{tabular}
    \caption{
        Examples of image pairs that are removed from the reference set prior to forming the dataset. 
        \label{fig:removedpairs}
    }
\end{figure}

Both data sources contain pairs of similar images, where the types of similarity cover the whole range, from exact duplicates to the same object category. 
To reduce the ambiguity of the ground-truth, we remove the most obvious images pairs from the reference set that correspond to the same instance but look like a case of edited copies.
There is a chicken-and-egg problem here because duplicates have to be identified to be removed, which is the purpose of the competition. 
We decided to use an in-house descriptor with a relatively narrow threshold to remove duplicates. 
Examples of removed pairs are shown in Figure~\ref{fig:removedpairs}.

%
%

%
%

\iffalse
\begin{figure} 
\begin{centering}
\includegraphics[width=\linewidth]{figs/isc_data.pdf}
\end{centering}
\caption{\label{fig:iscdata}
    Data generation diagram \matthijs{Not sure we want to go into that level of detail}.
}
\end{figure} 
\fi

\begin{figure}[t]
\begin{center}
\newcommand{\ige}[1]{\includegraphics[height=2.5cm]{figs/examples/#1}}
\newcommand{\ecom}[1]{#1}
\newcommand{\imcreds}[1]{\raisebox{\depth}{\scalebox{0.4}{\rotatebox{270}{Flickr / #1}}}}
\newcommand{\imcredsD}[1]{\raisebox{\depth}{\scalebox{0.4}{\rotatebox{270}{DFDC image #1}}}}
\begin{tabular}{cc@{}c}
\multicolumn{3}{c}{\ecom{text overlay + drawing}}\\
\ige{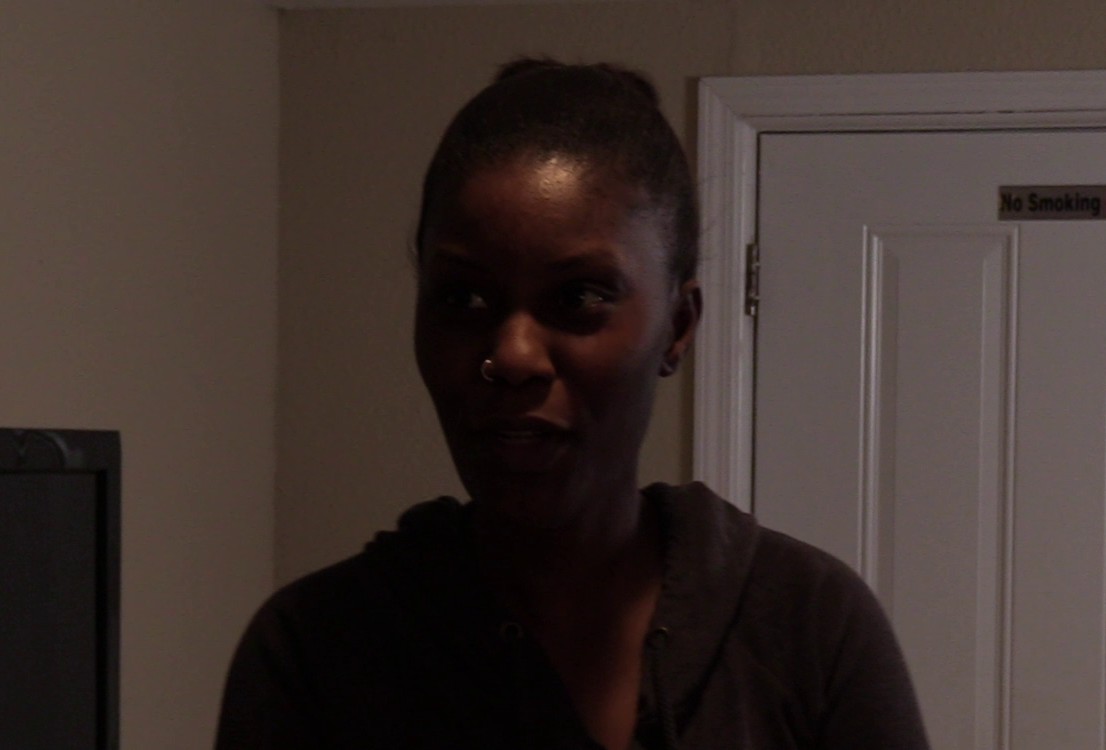} & \ige{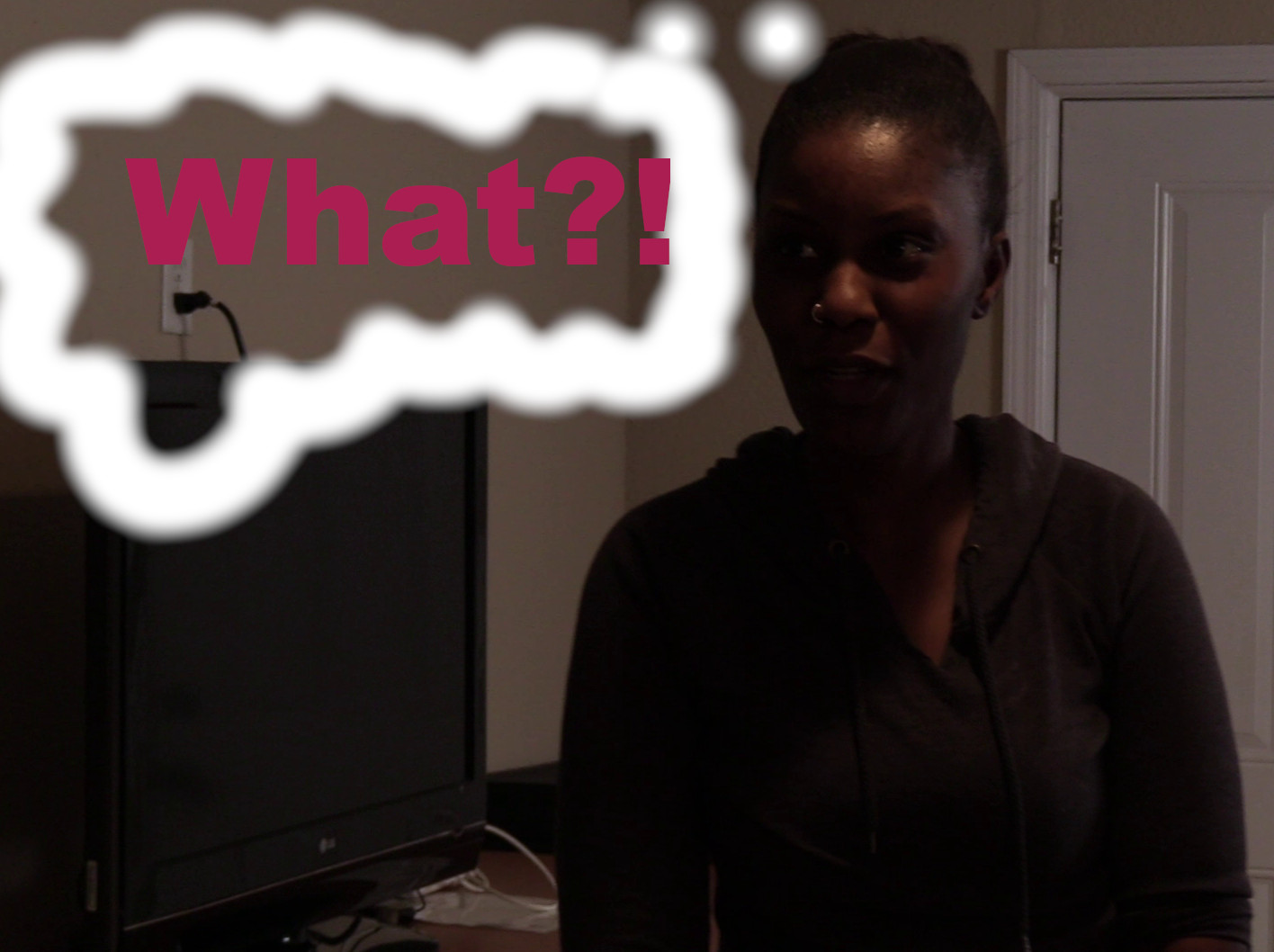} & \imcredsD{1757\_1}\\
\multicolumn{3}{c}{\ecom{paste salient part of image + pad + overlay text}}\\
\ige{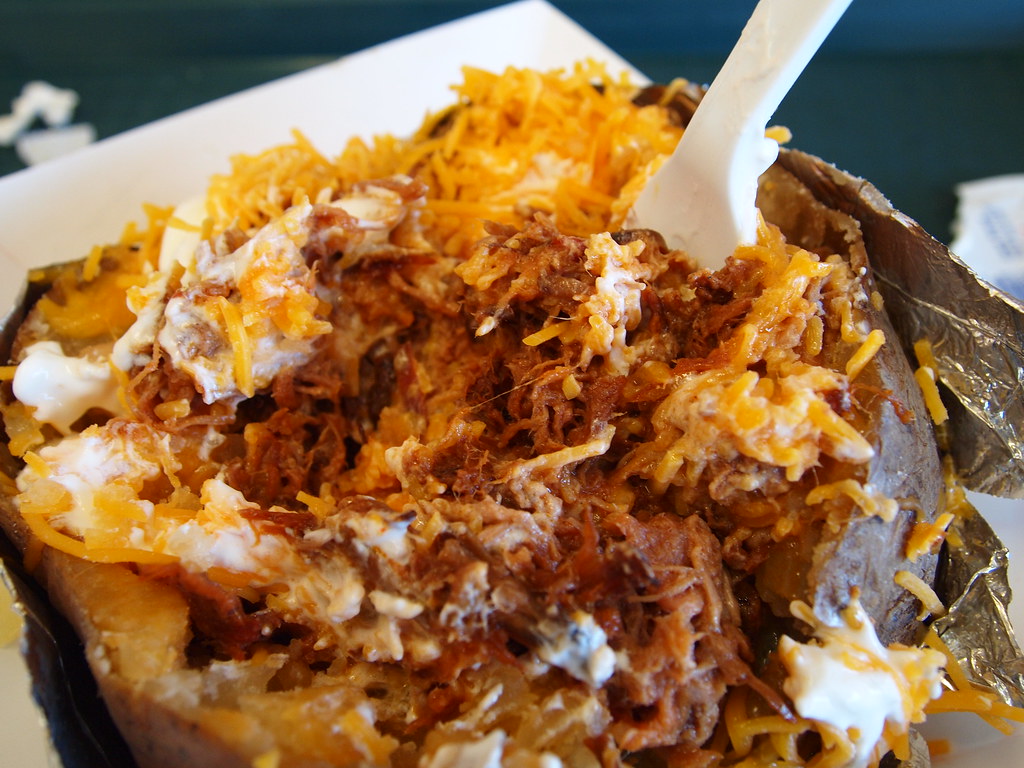} & \ige{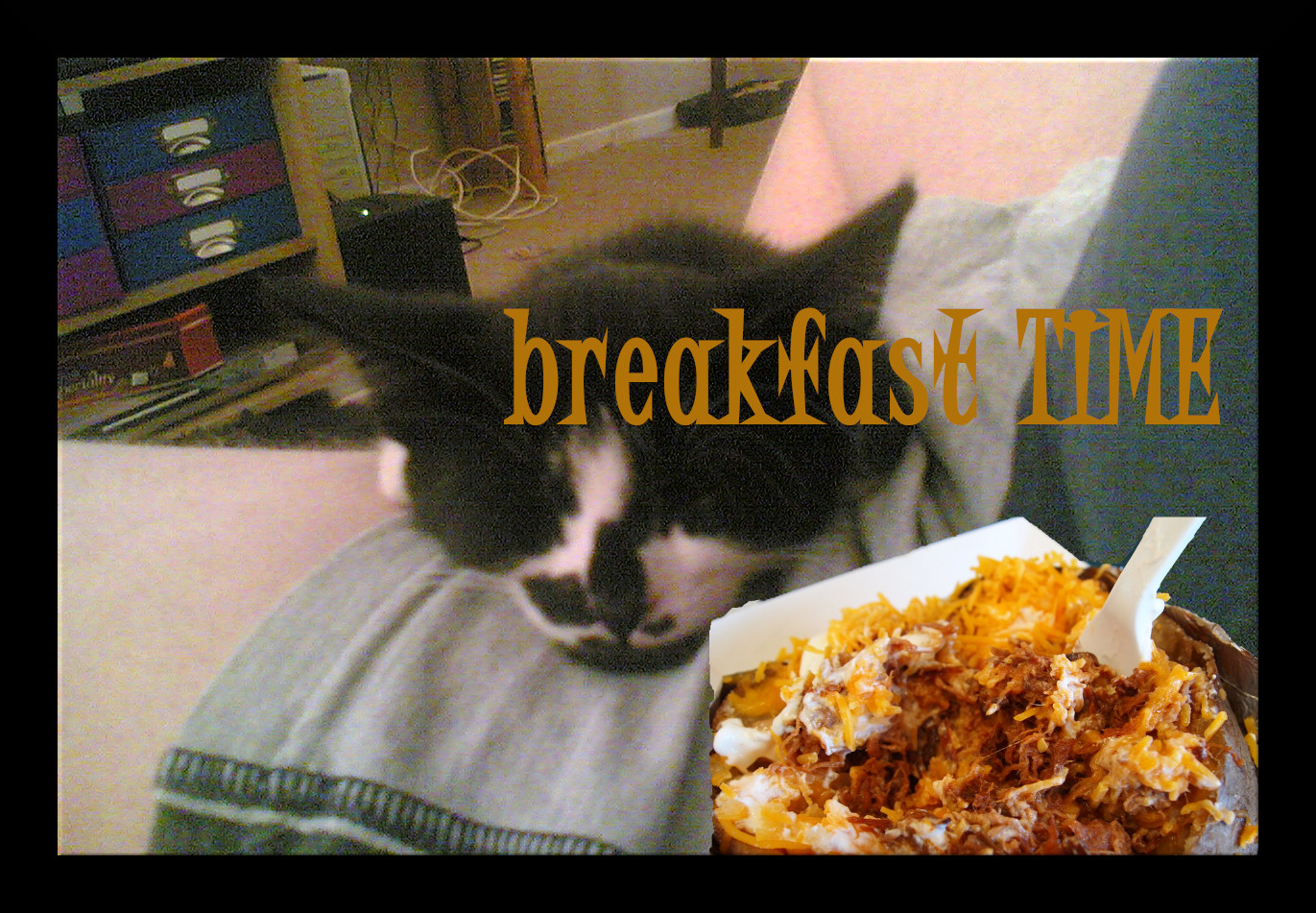} & \imcreds{Shreveport-Bossier} \\
\multicolumn{3}{c}{\ecom{handwriting + blur non-salient part of image}} \\
\ige{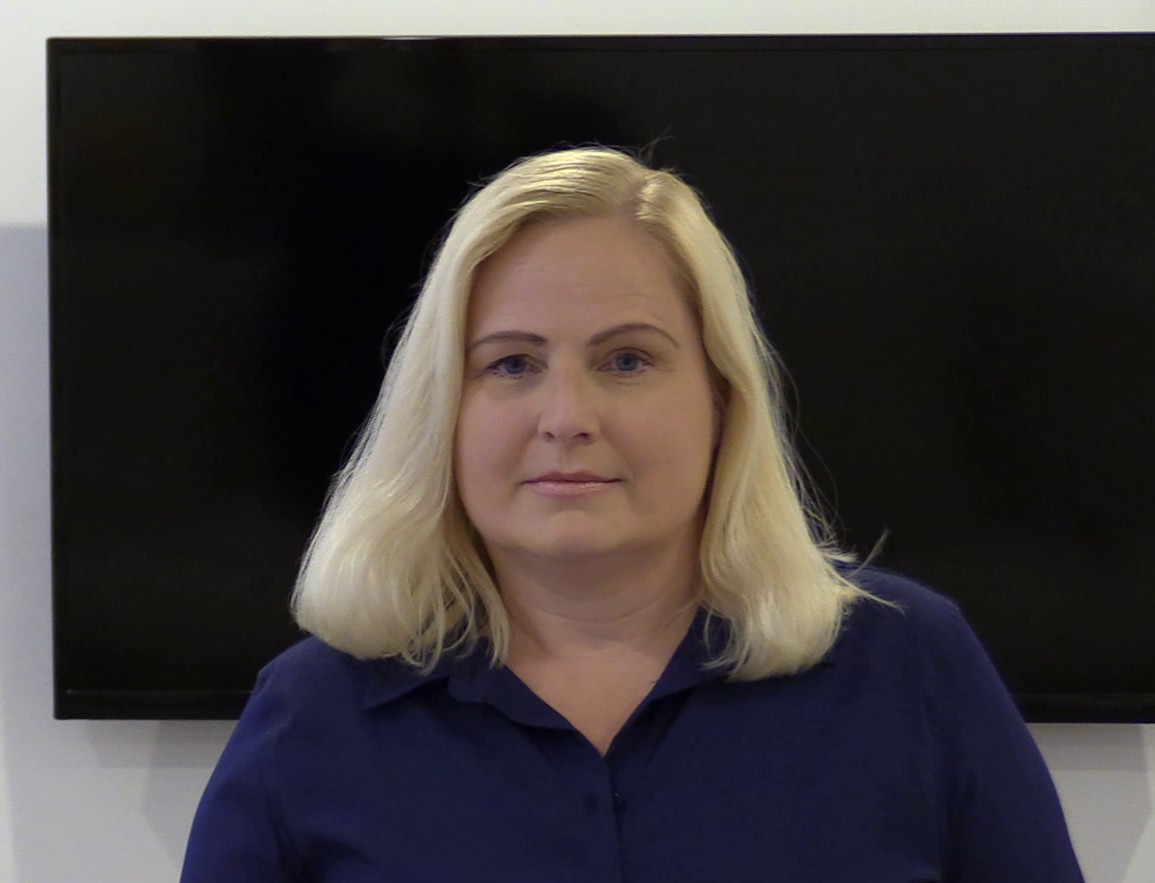} & \ige{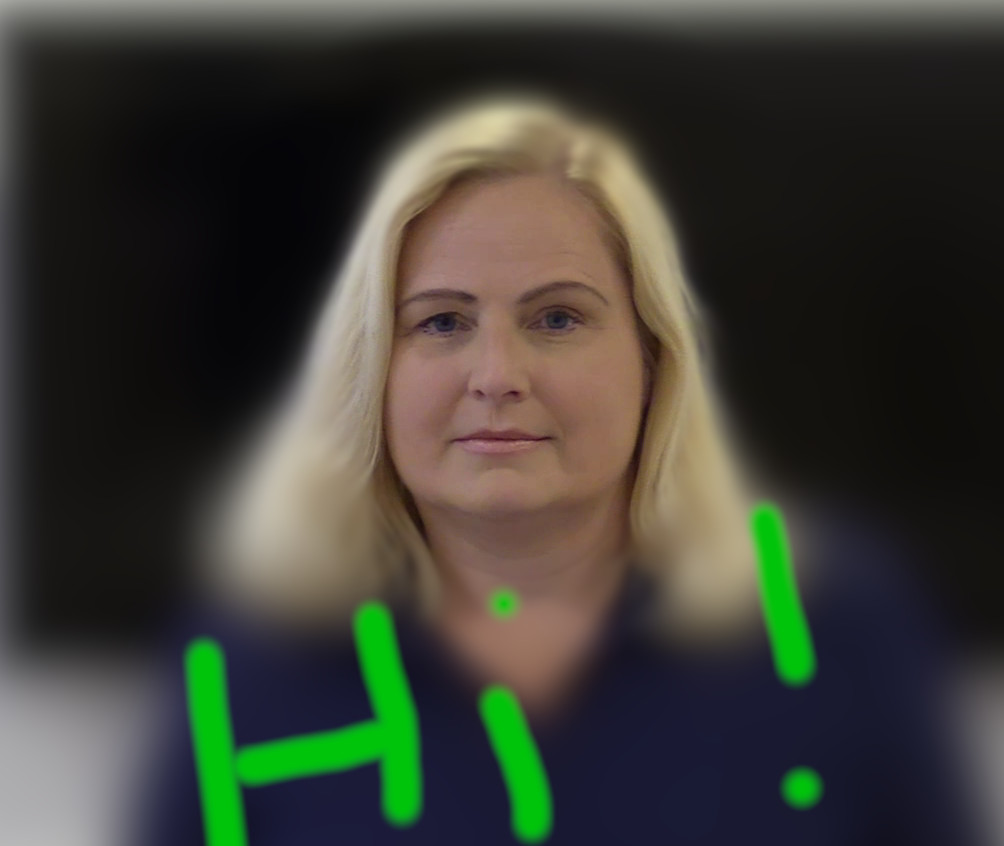} & \imcredsD{1128\_7} \\
\multicolumn{3}{c}{\ecom{perspective transform + circle salient part of image}} \\
\ige{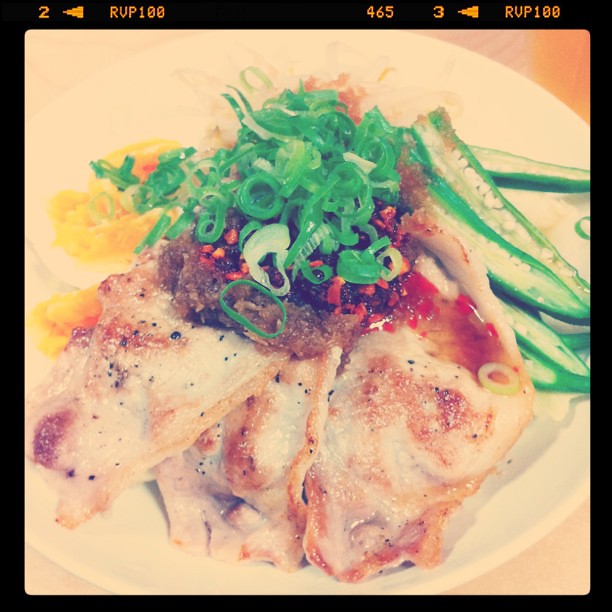} & \ige{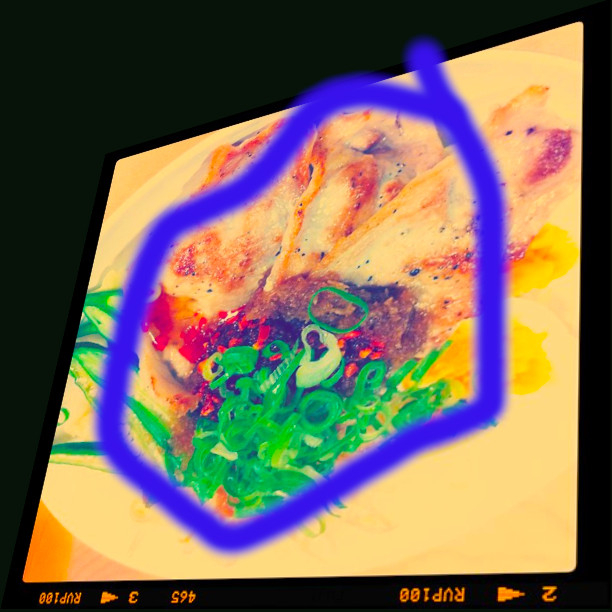} & \imcreds{senov} \\
\multicolumn{3}{c}{\ecom{paste salient part of background + pixelization}} \\
\ige{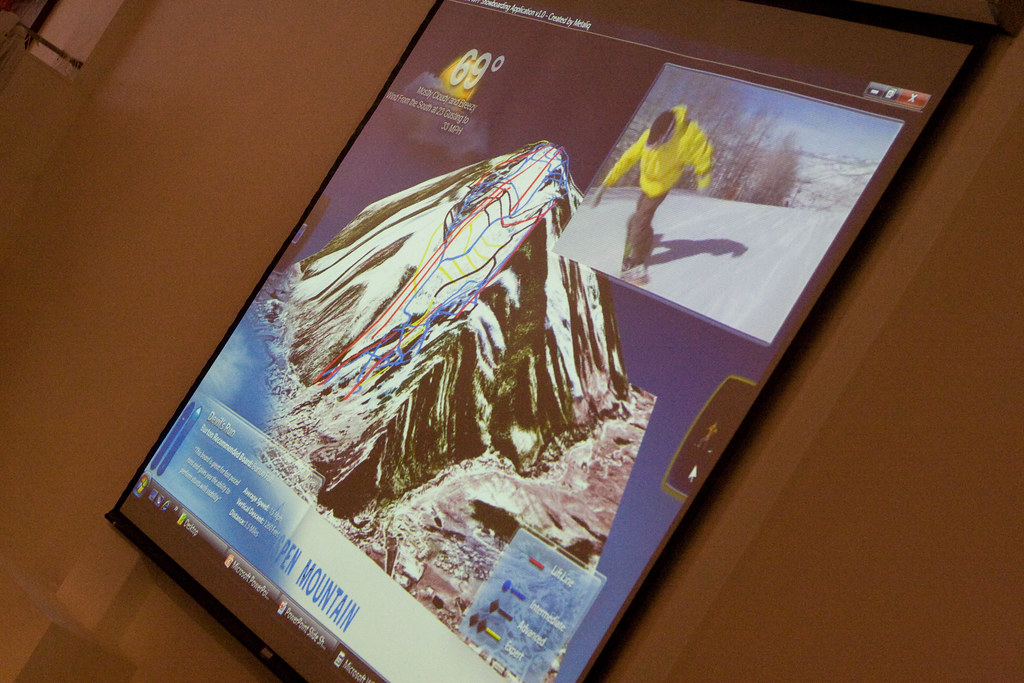} & \ige{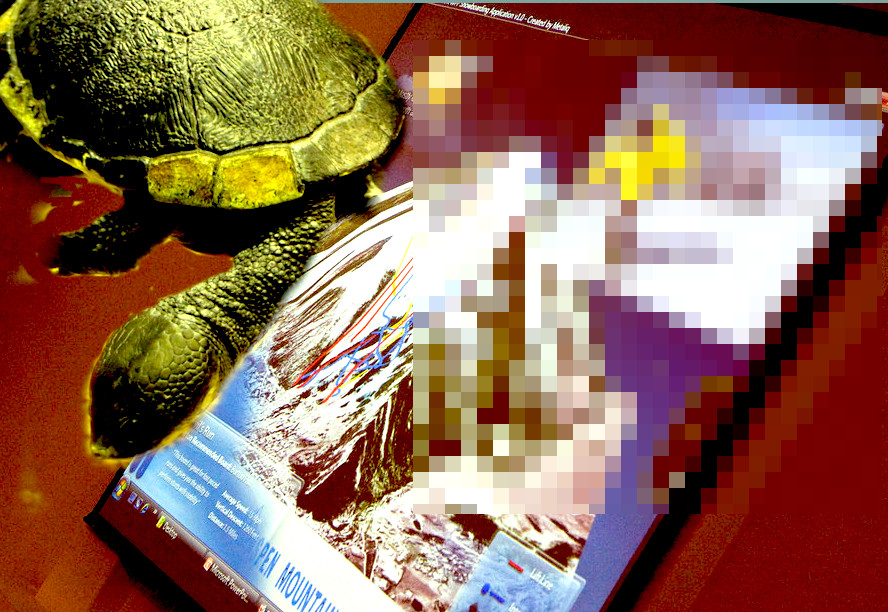} & \imcreds{roland} \\
\end{tabular}
\end{center}
\caption{\label{fig:transformedmanual}
Examples of source images and their manually transformed counterparts.
}
\end{figure}

\begin{figure*}[t]
\begin{center}
\newcommand{\ige}[1]{\includegraphics[height=2.5cm]{figs/examples/#1}}
\newcommand{\ecom}[1]{\multicolumn{3}{|c|}{#1}}
\newcommand{\imcreds}[1]{\raisebox{\depth}{\scalebox{0.4}{\rotatebox{270}{Flickr / #1}}}}
\newcommand{\imcredsD}[1]{\raisebox{\depth}{\scalebox{0.4}{\rotatebox{270}{DFDC image #1}}}}
\hspace*{-16pt}%
\begin{tabular}{|cc@{}c|cc@{}c|}
\hline
\ecom{rotate} & \ecom{screenshot overlay} \\
\ige{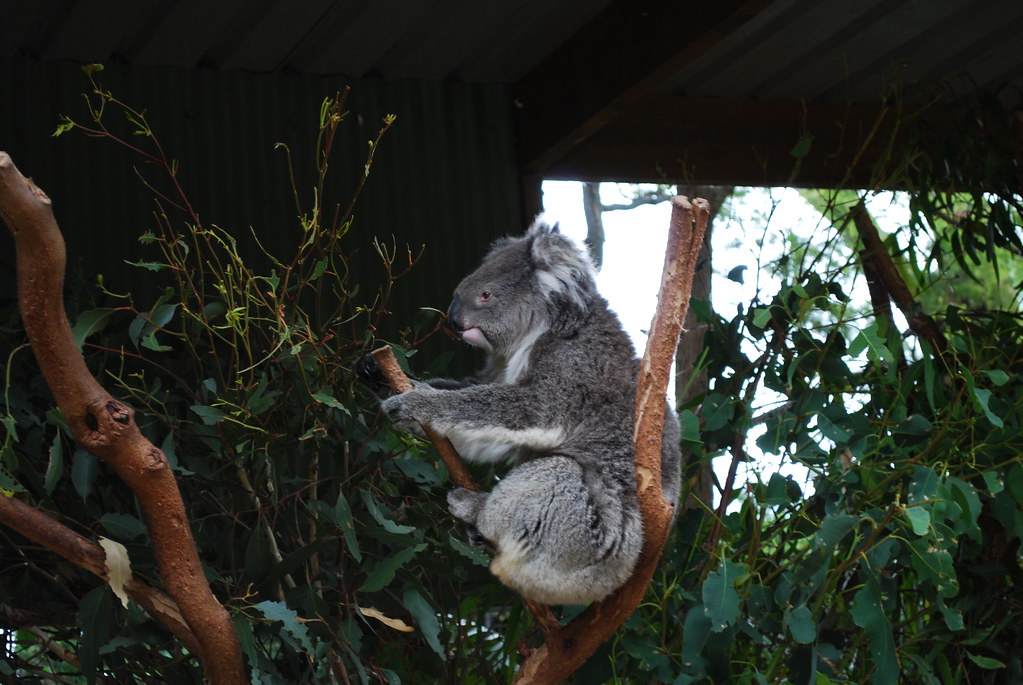} & \ige{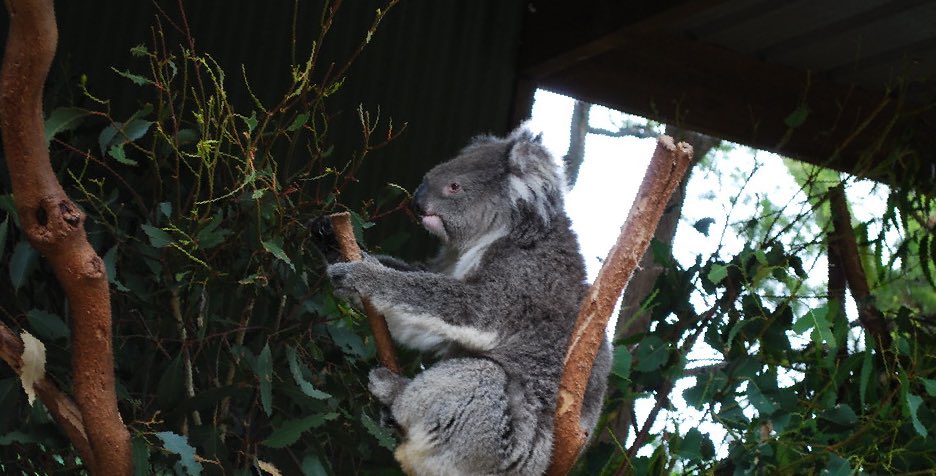} & \imcreds{jenster181} &
\ige{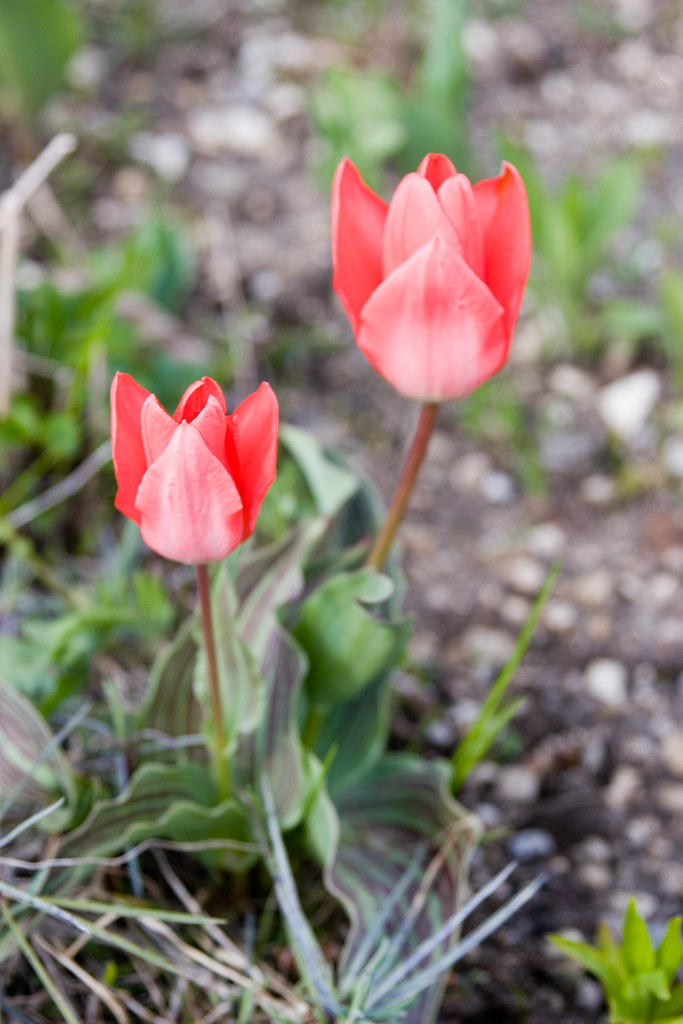} & \ige{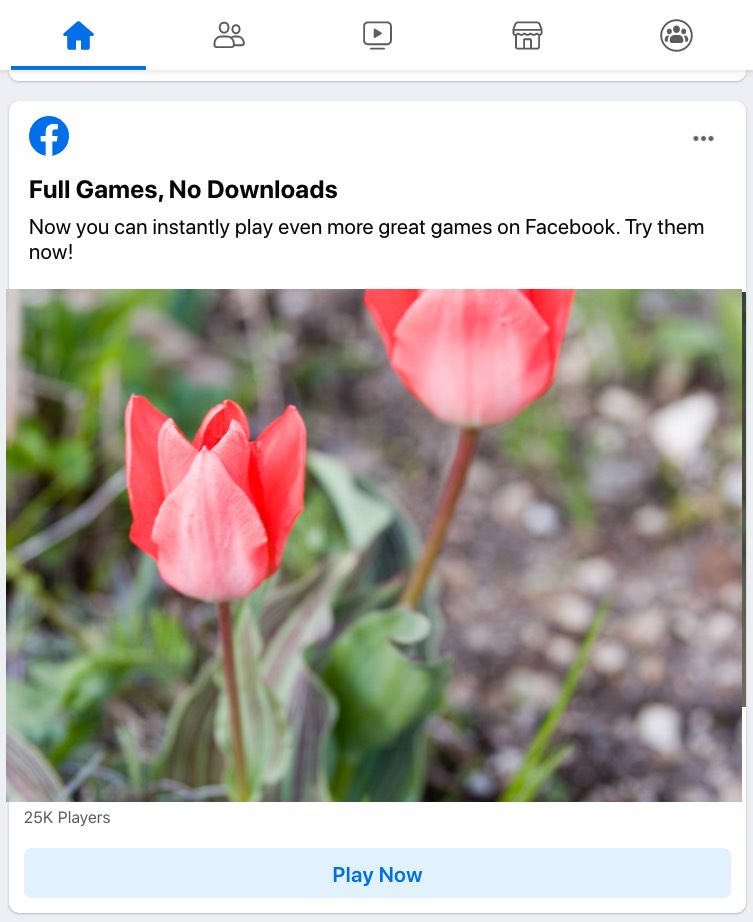} & \imcreds{Tilen Travnik} \\
\ecom{horizontal flip + IG filter} & 
\ecom{AR filter + text overlay} \\
\ige{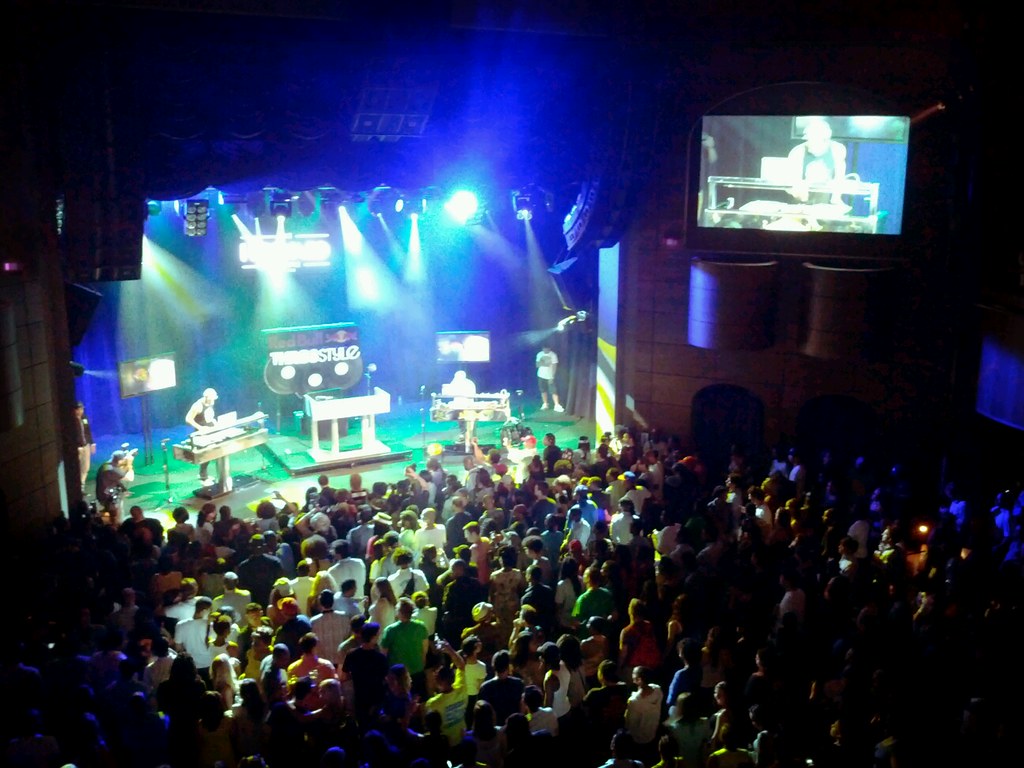} & \ige{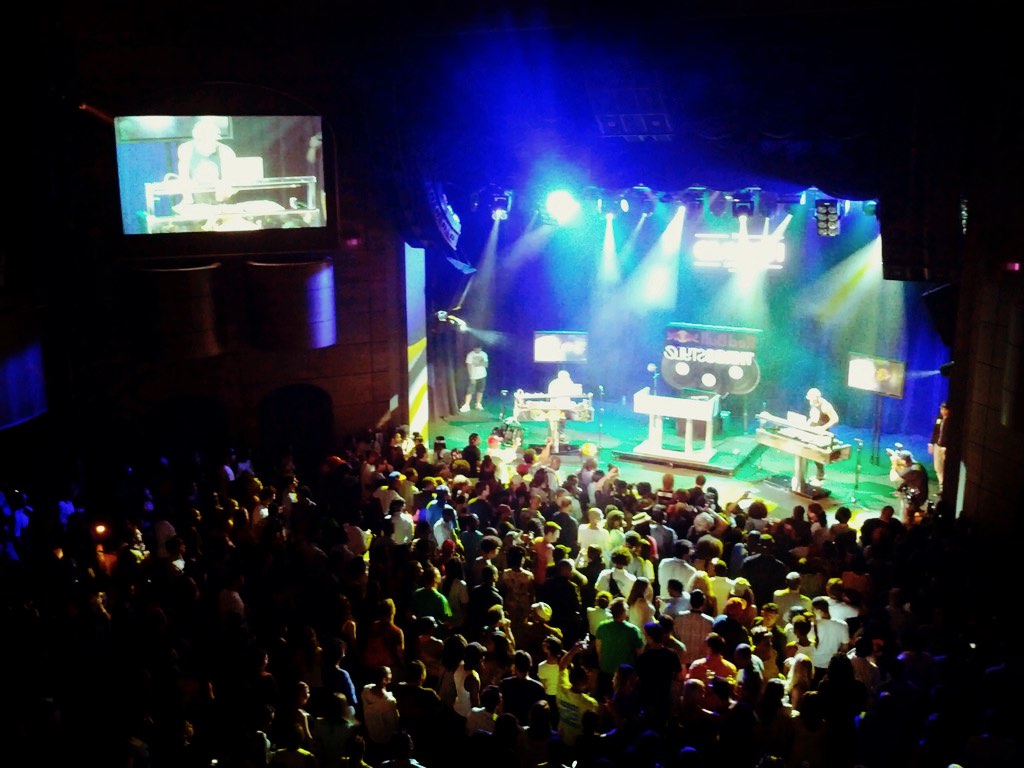} & \imcreds{zugaldia} &
\ige{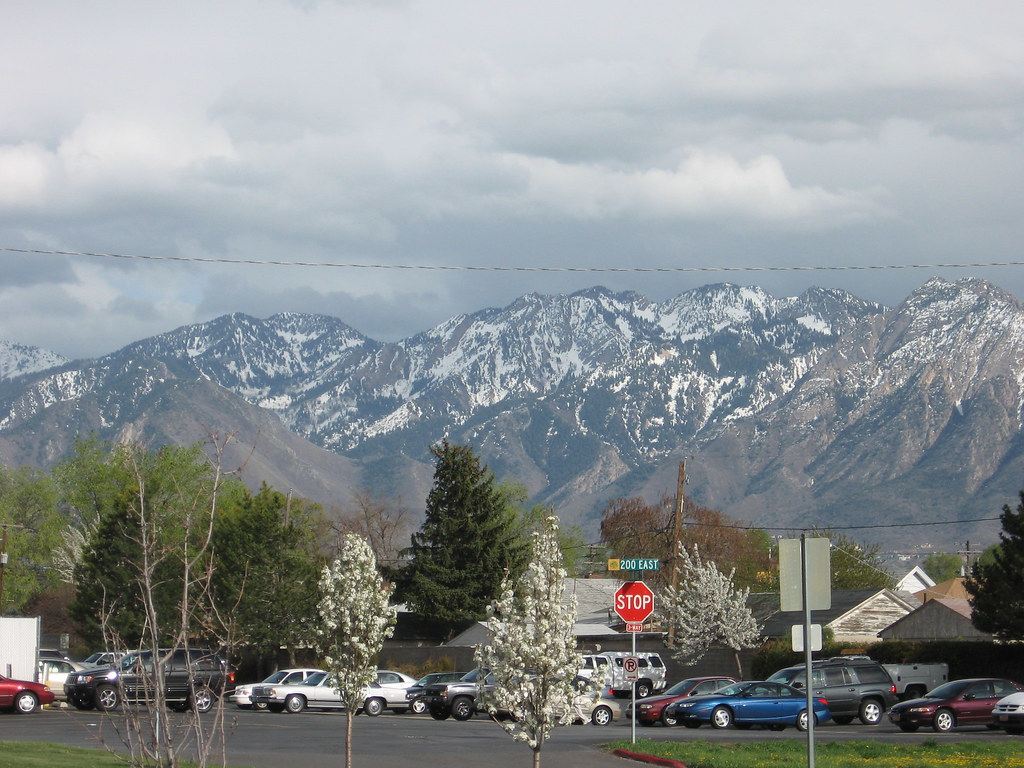} & \ige{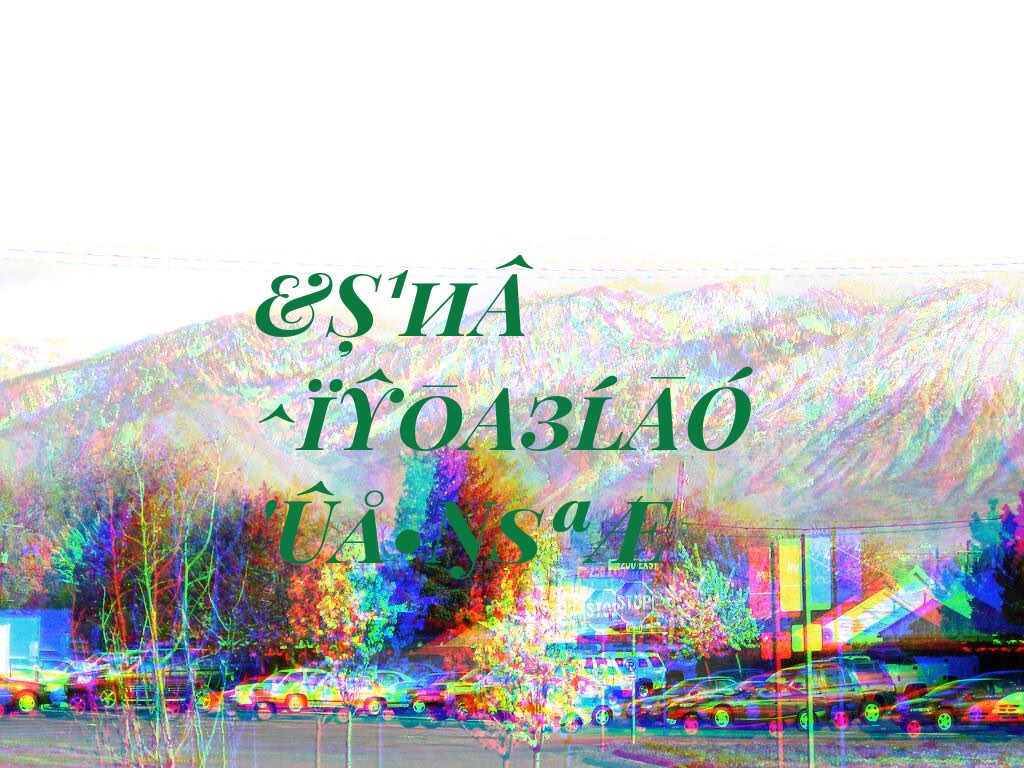} & \imcreds{Jim Moore} \\
\ecom{deepfake + text + AR filter} & 
\ecom{brightness + mask overlay + text overlay} \\
\ige{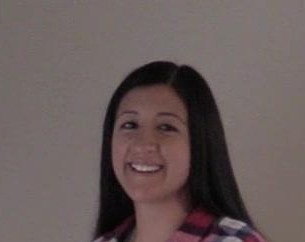} & \ige{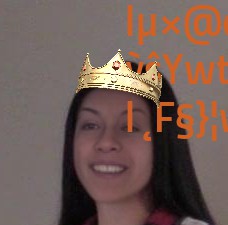} & \imcredsD{0989}& \ige{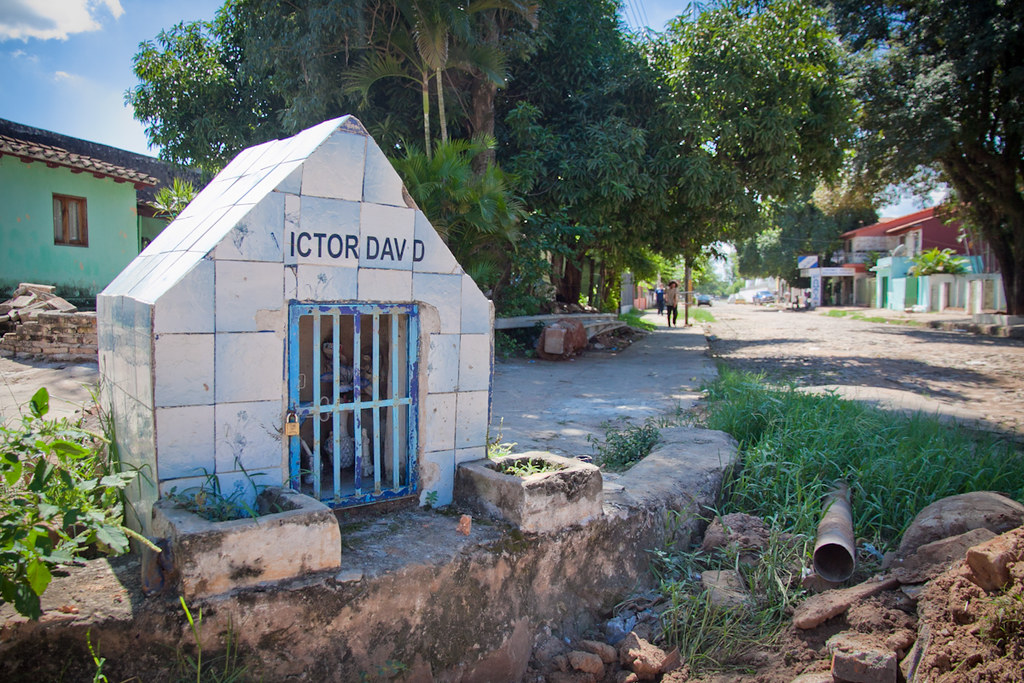} & \ige{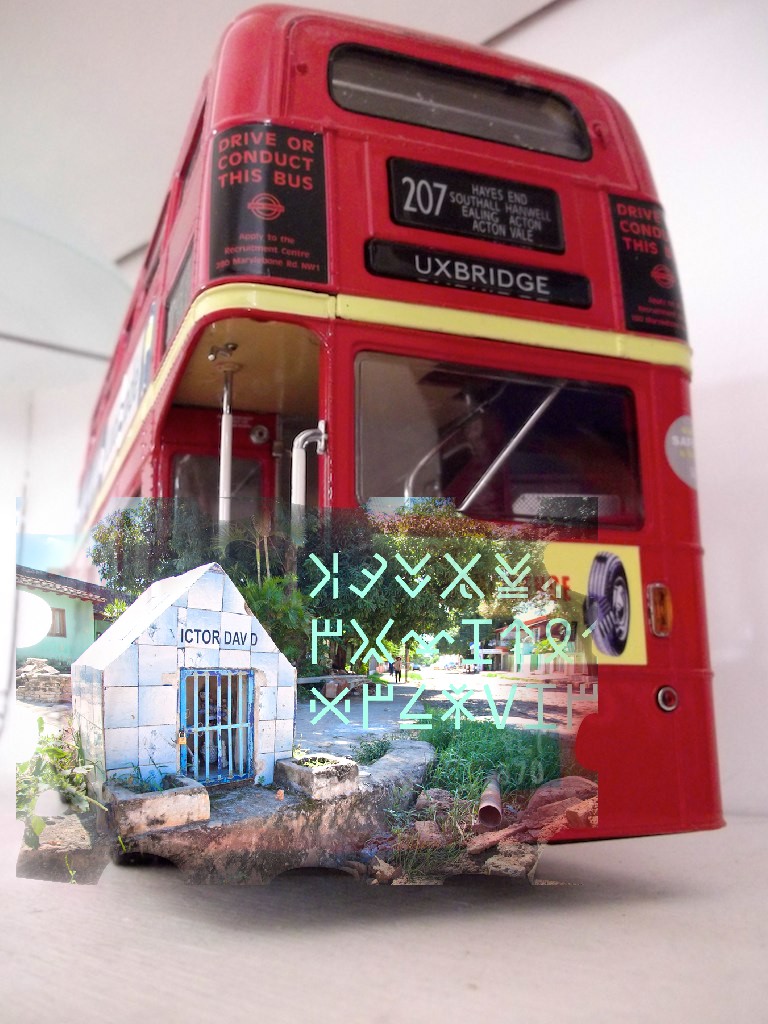} & \imcreds{Arcadiu\v{s}}\\
\ecom{blur + overlay text + overlay on background} & 
\ecom{saturation+pixelization+padding+emoji overlay} \\
\ige{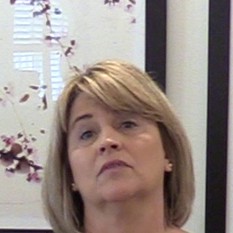} & \ige{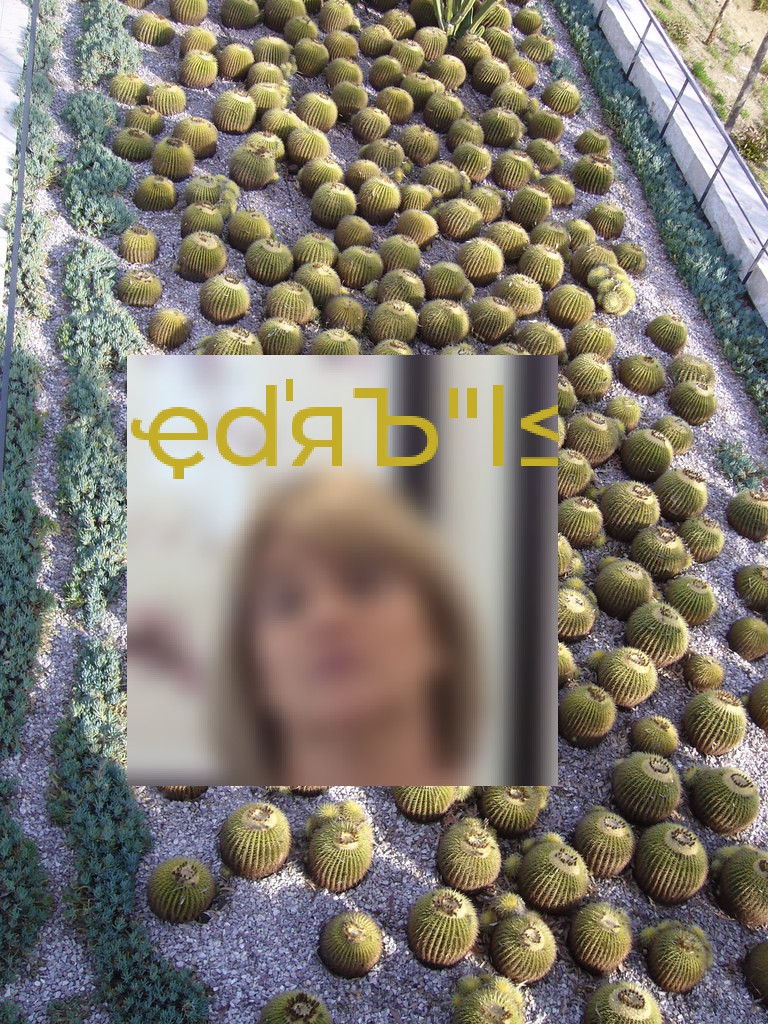} & \imcredsD{1812\_3} &
\ige{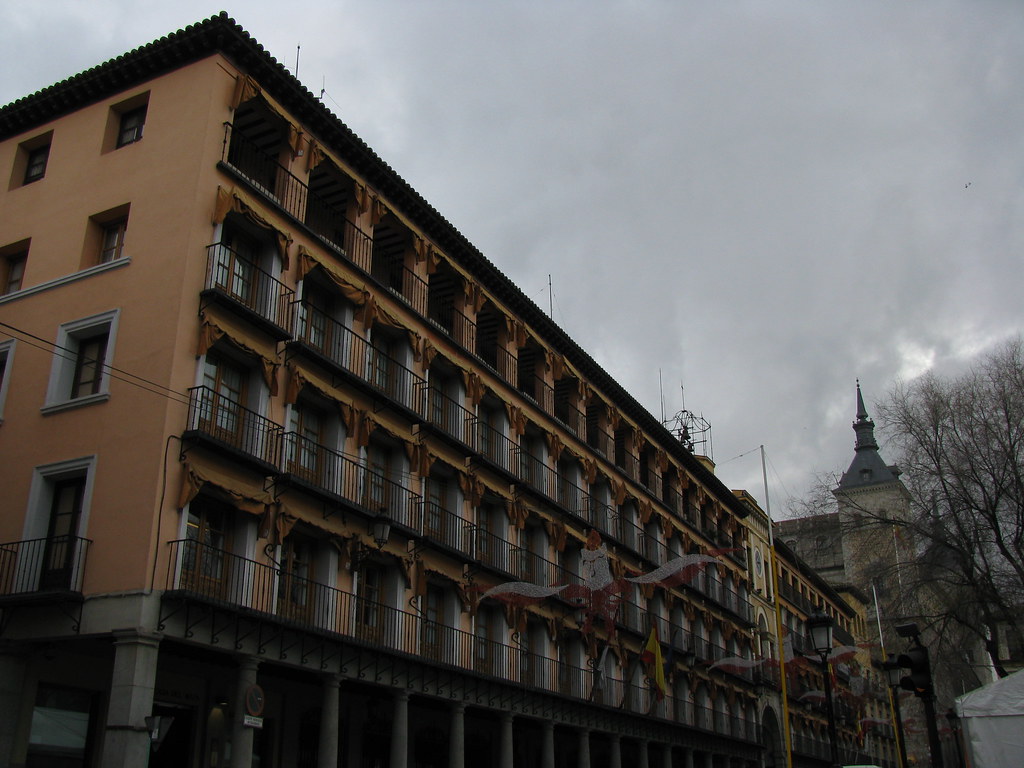} & \ige{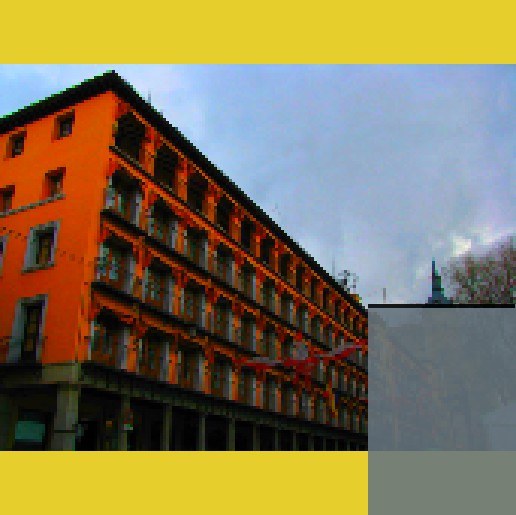} & \imcreds{Bogdan Migulski} \\
\ecom{grayscale+degrade quality+text overlay+image overlay} &
\ecom{enhance edges + mask overlay} \\
\ige{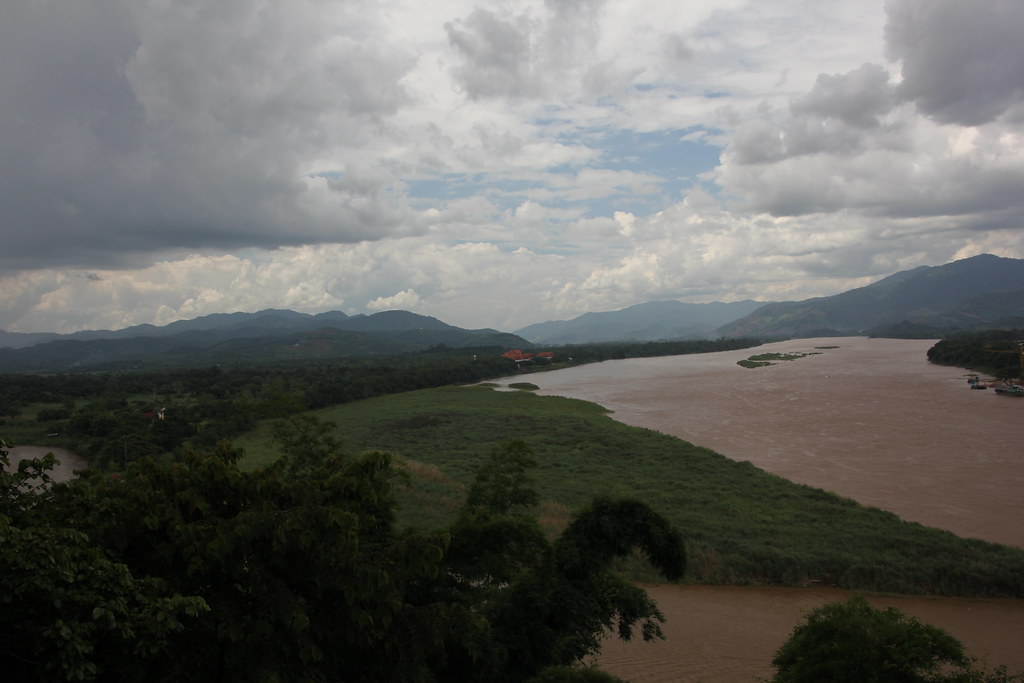} & \ige{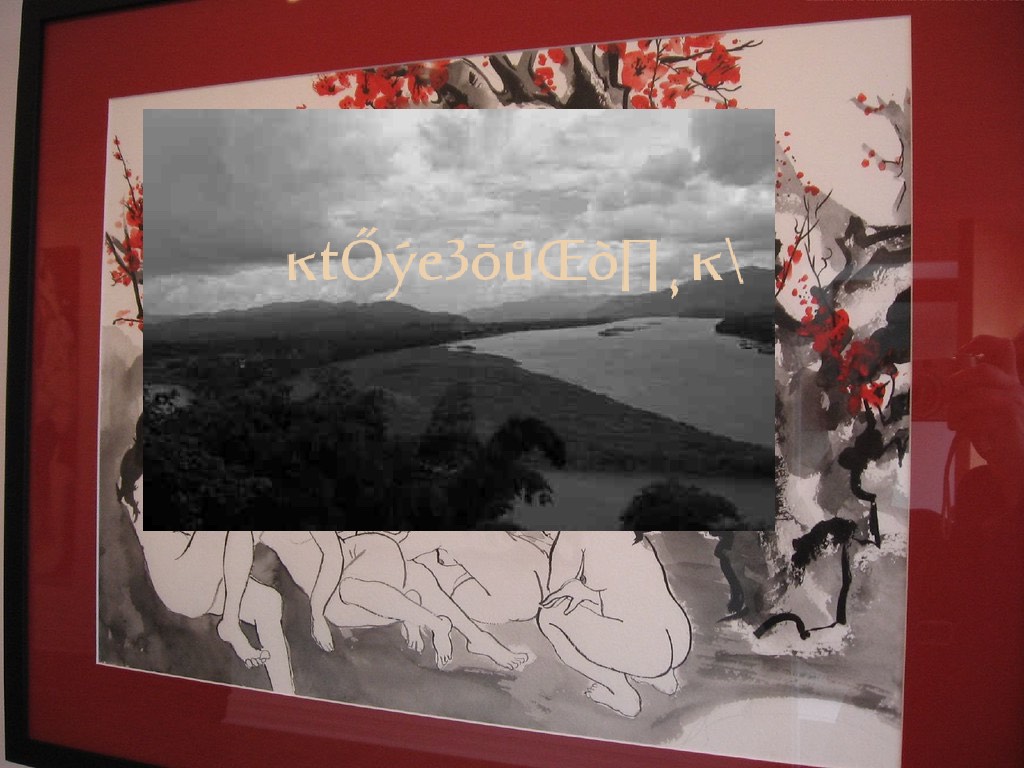} & \imcreds{Dave 79} & 
\ige{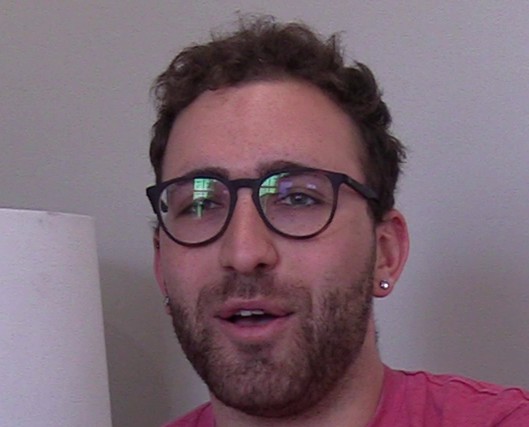} & \ige{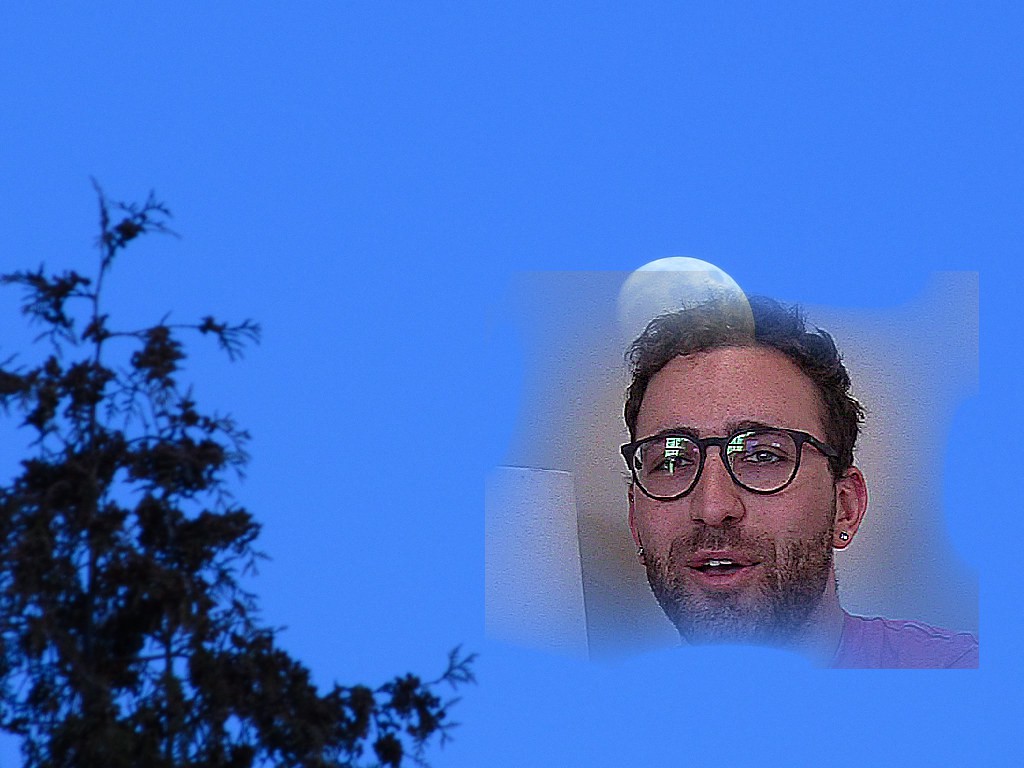}  & \imcredsD{3125\_13}\\
\hline
\end{tabular}
\end{center}
\caption{\label{fig:transformedautomatic}
    Example pairs of source images (left) and their automatically transformed counterparts (right). 
    The transformations are calibrated to match the range of difficulties encountered in a real-world setting.
}
\end{figure*}

\subsection{Transformations}

We do not use transformations of uniform difficulty, but instead target a wide range of difficulties, from easy near-exact duplicates to cases that are hard to assess by a human.
To ensure that the dataset is neither too easy nor too hard, during the collection of the dataset, we monitor its difficulty by running a set of existing baseline approaches on it.
The dataset is calibrated so that the performance of the different baselines ranges from 10 to 40\%. 
In this way, there are several hard transformations included so that the dataset can attract and encourage research effort for years before becoming obsolete.

Images are transformed via either manual (Figure~\ref{fig:transformedmanual}) or automatic (Figure~\ref{fig:transformedautomatic}) edits. 
All transformations start from a source image that has to be altered to a certain degree. 
Some transformations require a secondary image from the YFCC100M to serve as an overlay or a background. 
In that case, we make sure that this secondary image is not included in the reference set.

\paragraph{Manual transformations. }
Manual transformations are performed by external image editors with a photo editing software. 
We chose GIMP as the software because it is free, available on all types of computers, and has a very large toolbox of image manipulation functions.
The instructions provided to the editors are as follows: 
\begin{itemize}
    \setlength\itemsep{0.05em}
    \item Editors are provided with one source image and optionally a secondary image that they can use to make collages
    \item Each edit should use 2 to 5 different tools in GIMP
    \item All editing tools are acceptable including geometrical deformations, color transformations, brush strokes, image filters, etc.
    \item Diversity is important; tools, as well as their parameters, should be changed as often as possible
    \item Once editors are familiarized with the task, the mean time to edit an image should be around 3 minutes
    \item The result should be distinctive enough so that the source image is still recognizable to the editor   
\end{itemize}

\paragraph{Automatic transformations. }
Automatic transformations are applied using AugLy~\cite{augly}, a data augmentations library created at Facebook AI. 
The automatic transformations are classified into the following broad categories:

\begin{itemize}
    \setlength\itemsep{0.05em}
    \item Overlays: text and emoji
    \item Color transformations: changes in brightness or saturation, grayscale, Instagram color filters and Augmented Reality effects
    \item Pixel-level transformations:
    blur, color palette with dithering, JPEG encoding, edge enhance, pixelization, pixel shuffling
    \item Spatial transformations:
    crop, rotation, horizontal flip, padding, aspect ratio change, perspective transform, overlay onto background image
    \item  embedding the image into the GUI of a social network application
\end{itemize}

Some images have only one automatic transformation applied, while some others have multiple chained together.
The transformations and parameters are picked at random.
The distribution and strength of the transformations is roughly based on what is witnessed in the production setting at Facebook, but with a stronger emphasis on difficult transformations.

In the second phase of the competition we added other transformations and made the existing ones more difficult. 
Appendix~\ref{sec:moretransfo} details these transformations. 

\subsection{Dataset structure}

\OURD consists of the following parts: the reference image set, the training image set, and two query image sets. We create two query sets in order to mimic the real-world setting, where there is a drift over time in the applied transformations, by adding a few additional transformations to the second set.
This setting is aligned with the two phases of the competition (see Section~\ref{sec:rules}) and is intended to penalize methods that overfit to a fixed set of transformations, \ie the ones in Phase I. 

\begin{itemize} 
\item
    Reference image set: 1 million images without any transformation. 
    The YFCC100M dataset is publicly available, but we will re-host the chosen subset.
\item 
    Training image set: 1~million images collected in the same way as the reference set. 
    The intended use is for all kinds of training tasks, especially the ones that depend on the data distribution of the reference images, such as model training via data augmentation, scores normalization, and PCA training.
\item 
    Development query image set (Phase I): 10,000 images from the reference set mixed with 40,000 distractor images that are not part of the reference set,  that have been edited in various ways. 
    Distractor image queries have no matching counterpart in the set of 1 million reference images. 
\item Test query image set (Phase II): 10,000 images from the reference set mixed with 40,000 distractor images. All these images are transformed in various ways to form the query images. This set, compared to the one of Phase I, is generated by including a few additional transformations and the transformation parameters are slightly modified.

\end{itemize}

\subsection{Evaluation metric}
\label{sec:metric}

The output of an image copy detection algorithm is a set of pairs, where each pair is formed by a query image and a candidate source image from the reference set, and is accompanied by a confidence value. 
Not all queries need to be part of this list as there are many distractor queries. 
We use micro Average Precision to measure performance, which is described in the following and only after precision and recall measures are discussed.

\begin{figure}
    \centering
    \includegraphics[width=\linewidth]{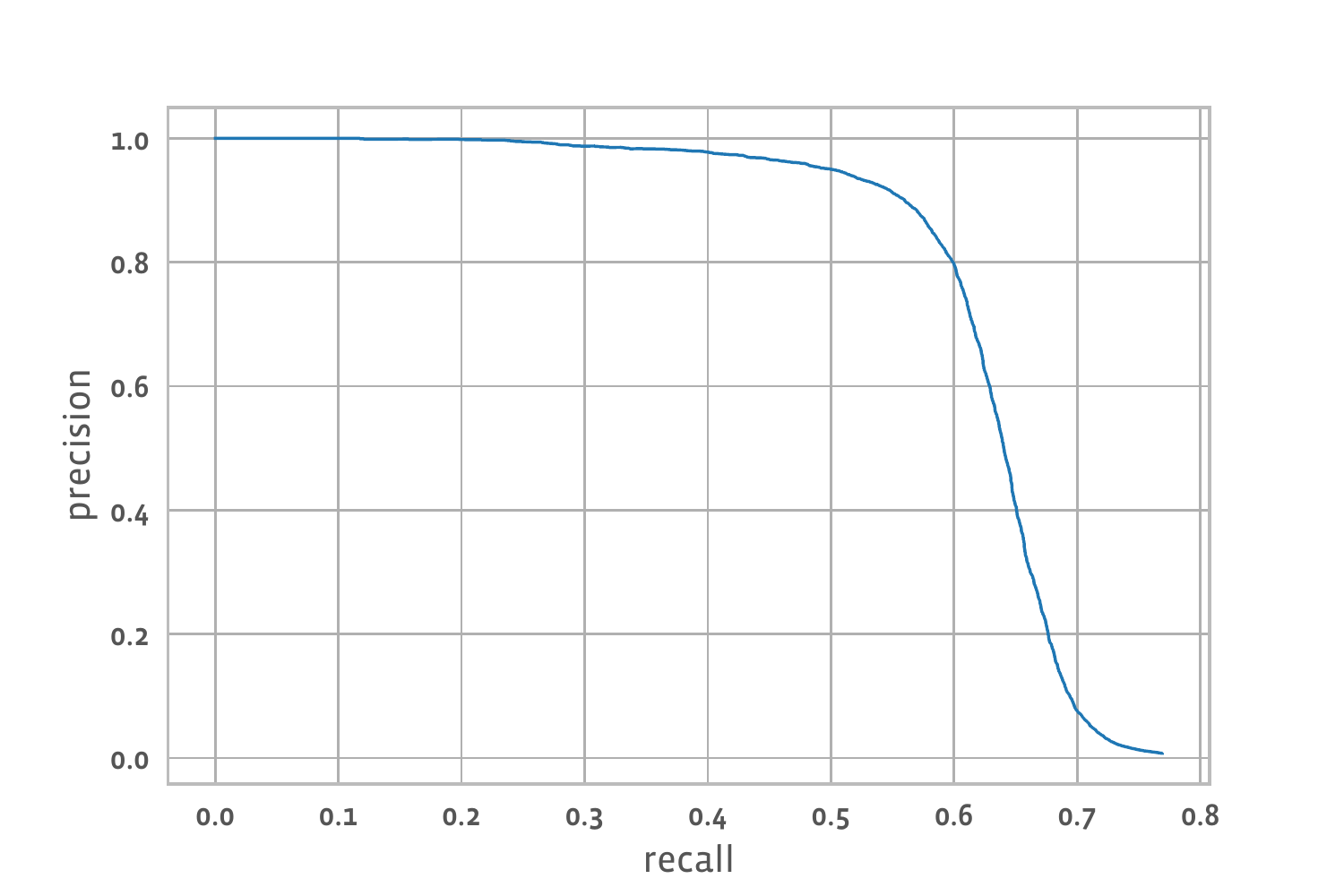}
    \caption{
        Example precision-recall plot for copy detection with 50,000 queries and a reference set of 1 million images. 
        The total number of queries that are edited copies of images in the reference set  is 10,000. 
        A quick read of the plot shows that one threshold value on the scores obtains a recall of 60\% for 80\% precision.
        This means that 6,000 copies are properly detected (true positives) but 1,500 copies are incorrectly detected (false positives).
        The corresponding metric that we use for the benchmark is the area under the curve, $\mu AP=0.62781$ in this case.
    }
    \label{fig:prplot}
\end{figure}

\paragraph{Precision and recall.}

Detected pairs for all queries are considered simultaneously.
Setting a threshold on the confidence amounts to taking a hard decision whether there is a detection or not.
Then, the precision and recall of the detected pairs are evaluated according to the ground-truth. 
Note that detected pairs for distractor queries are always wrong detections as the corresponding source image is not part of the reference set. 
By varying the threshold, one can adjust the trade-off between precision and recall, as in Figure~\ref{fig:prplot}. 

Precision and recall measures can be extrapolated using basic arithmetic to larger reference sets than do not have additional positive pairs of images. 
For example, extrapolating precision equal to 0.8 and recall equal 0.6 measured on our 1~million image reference set to 10~million images, then, for a given threshold, the recall does not change  as there are no additional positives, but the number of false positives increases by a factor 10. Therefore, the new precision drops to around 0.29.
Incidentally, the low precision explains why at a large scale, image copy detection algorithms operate mostly in a high-precision and low-recall regime.

\paragraph{micro Average precision.}
We measure the overall performance, for all possible thresholds, by the average precision.
It is equivalent to the area under the precision-recall curve when all pairs are taken into account and is also known as micro-AP ($\mu AP$).
This measure has been used for instance recognition~\cite{perronnin2009family,google2019landmarkschallenge}.
It is computed as
\vspace{-3pt}
\begin{equation}
\mu AP = \sum_{i=1}^{N} p(i)\Delta r(i) \in [0, 1],
\vspace{-3pt}
\end{equation}
where $p(i)$ is the precision at position $i$ of the sorted list of pairs, $\Delta r(i)$ is the difference of recall between position $i$ and $i-1$, and $N$ is the total number of returned pairs for all queries. 
Note that any detected pair for a distractor query will decrease the AP score.

Thus, all queries are evaluated jointly by merging returned pairs for all queries and sorting them according to confidence; a single precision-recall curve is generated.
This is different than mean-Average-Precision (mAP), also known as macro-AP~\cite{perronnin2009family}, where AP is computed per query separately and then averaged over all queries. Confidence values matter with micro-AP, while only the ranking per query matters with macro-AP. The former better reflects the objectives of image copy detection, while the latter is typically used in retrieval tasks.

\subsection{Discussion}

The design of the dataset aims at striking a tradeoff between realism and ease-of-use as a computer vision benchmark. 
This simplification introduces biases in the dataset that we discuss here. 

\paragraph{Biases.}

A good fraction of social media images contain \emph{text and faces}.
Specialized techniques that focus on these features, such as OCR matching and face recognition, might be helpful on small benchmark datasets, but would not advance copy detection in the general case.
We design the benchmark to be insensitive to these specialized recognition techniques in the following ways. 
Firstly, text is used often as a transformation; relying on text recognition will yield false positives. Secondly, we include face-specific transformations that render the face unrecognizable or change the person's identity, \ie ``deep fakes''~\cite{DFDC2020}.

Another bias is that \emph{each source image is transformed only once}, which does not follow a real-world setting where user uploads can become viral. This design choice maintains the sensitivity of the benchmark and prevents from determining copy detection performance primarily by the performance on the few source images that are copied most of the times.

Another bias is that the \emph{source images are not transformed}. This is realistic in a copyright context where we have access to the original images, but less realistic for other cases. However, it simplifies the setup. 

The final bias is that the dataset focuses on too strong transforms.
This focus will favor the emergence of expensive and complex approaches that are not required to handle the vast majority of production use cases. 
The rationale for this is that we often observe that it is useful to have a ``topline'' method in terms of accuracy.
Other methods can then be derived from it by degrading its parameters.

\paragraph{Annotation errors.}
We do our best to provide a clean dataset and ground-truth. However, since the dataset is partially generated automatically and not validated manually, there could be annotation errors.
These errors could be matching images in the reference set or query images that are not recognizable because the source content has disappeared completely.
We do believe that the impact of these errors is low and most importantly will not incur a bias between different methods.

\section{Baselines}
\label{sec:baselines}

We report results with a number of baseline methods. Popular and successful image retrieval methods were selected and adapted from publicly available code.

\begin{table*}[t]
    \centering
  \begin{tabular}{lrrrrrr}
    \hline
    method & dimension & w/o norm. & with norm. & time (ms) & hardware\\
    \hline
    GIST 960 dim                     & 960  & 14.42 & --- & 0.55  & CPU, 2.2 GHz, 40 threads \\
    GIST PCA 256                     & 256  & 15.56 & --- & 0.55  & CPU, 2.2 GHz, 40 threads \\
    Multigrain 1500 dim              & 1500 & 16.47 & 36.42 & 23 & Tesla V100 \\
    HOW+ASMK                         & N/A  &   17.32        &  37.15 & 150 & Tesla P-100\\
    \hline
    \end{tabular}
    \caption{
    Performance measured with $\mu AP$ for the baseline methods on the development query set of \OURD. 
    For methods that rely only on pairwise global descriptor matching, we indicate the descriptor dimension -- if this dimension is below 256 they are eligible to submit to track 2 of the competition.
    The reported times are feature extraction times per image on the specified hardware. 
    This excludes the matching phase. 
    \label{tab:baselineresults}
}
\end{table*}

\subsection{Representative methods}

We choose three baseline methods from existing state of the art, and adapted them minimally for the challenge. 
We present by increasing complexities.

\paragraph{GIST.} 
The GIST~\cite{oliva2001modeling} is a very simple descriptor that is attractive because it is cheap to extract, works on low-resolution images and does not require any training. 
It was evaluated in~\cite{Douze2009EvaluationOG} for image copy detection and found to work reasonably well for JPEG encoding transformations and small crops. 

It is extracted by (1) resizing the image to 32$\times$32 pixels (2) splitting the image in 3 channels and a 4$\times$4 grid of cells, and (3) building a gradient orientation histogram in each cell. 
The output is a 960-dimensional vector.
This descriptor can be matched and used for the competition's track 1. 
For Track 2, the descriptor has to be reduced to 256 dimensions (see Section~\ref{sec:tracks}). 
We do this using a PCA, followed by L2 normalization. 
Note that searching in the full dimension or applying PCA with whitening does not benefit the GIST descriptor.

\paragraph{Multigrain.}
The multigrain descriptor~\cite{berman2019multigrain} is an early example of global image embedding based largely on data augmentation. 
Multigrain is a Resnet50 model trained on Imagenet. 
In addition to the standard classification head, the multigrain model has an additional head that does a generalized max-pooling (GeM)~\cite{radenovic2018fine} of the last activation map to generate an image embedding. 
The input resolution is set to 512 pixels in the largest dimension and the GeM parameter $p=7$. 
The image embedding enters a contrastive loss term that forces multiple data augmentations of the image to map to the same embedding. 
The GeM embedding has 2048 dimensions. 
It is reduced by PCA and whitening to 1500 dimensions and normalized. 
The resulting score is the inner product of the query and reference embeddings.

\paragraph{HOW deep local descriptors with ASMK.} In HOW~\cite{tolias2020learning}, individual activations of the CNN are treated as local descriptors. Even though the activations are treated as local descriptors at inference time, the network is trained with image level annotations only. During inference, the local descriptors are vector-quantized into visual words. Each local descriptor is assigned to a visual word and has a signature relative to that visual word. An image is then represented as a set of visual words, each with a signature aggregated for all descriptors associated with that word. These aggregated signatures are then binarized, following the ASMK approach~\cite{tolias2013aggregate}. An inverted-file structure is used for efficient similarity estimation between a query image and each image of the reference set. We use the public source code and the publicly available network (ResNet50) that is trained on a large training set of popular landmarks and buildings. We use default values for all hyper-parameters for inference.

\paragraph{Similarity normalization}
The baseline approaches provide a way to estimate the image-to-image visual similarity (confidence value); query image to reference image in particular, for all reference images. It is important for copy detection, as also captured by the evaluation metric, that the similarity values are comparable across different queries. 
We consider the similarity normalization introduced in~\cite{jegou2011exploiting}, which is later re-used at training time for word embeddings under the name CSLS~\cite{conneau2017word}.
Let $s(q,r)$ be the similarity between query image $q$ and reference image $r$. The normalized similarity is given by subtracting the similarity score a known non-matching image from the training set defined as
\begin{equation}
    \hat{s}(q, r) = s(q, r) - \beta s(q,t_n)
\end{equation}
where $t_n$ is the $n$-th most similar image to the query image from the training set. 
Note that since the normalization factor is estimated on the training set and not on  the reference set the final confidence score is independent of other reference images. It is also independent of other query images. The requirements are part of the rules for the research competition as described in Section~\ref{sec:rules}.

\subsection{Baseline results}

The matching results on \OURD for the three baseline methods are reported in Table~\ref{tab:baselineresults} without and with similarity normalization. The 500k pairs with the highest confidence score are used to estimate the evaluation metric. 
Hyper-parameters ($\beta$,$n$) for the normalization are set equal to $(0.5, 10)$ for GIST and to $(2,10)$ for the other two methods.

Computing GIST descriptors on 1~million images takes less than 20~minutes on a fast machine. 
Hence, the achieved performance is a good baseline for a cheap and relatively inaccurate descriptor.
It detects the lightest image edits in the collection.
Interestingly, reducing the dimension of the vector to 256 by PCA improves the results, so using the full 960-dimensional descriptor has no advantage.
Multigrain performs better than GIST, while HOW+ASMK offers a small performance increase with a significantly larger query computational cost though.

From the results, it is obvious that the score normalization has a strong impact on the $\mu AP$. 
This is because the methods that we used as baselines are mostly developed for ranking results to individual queries. 
We expect that techniques that explicitly optimize the estimation will make that normalization unnecessary.

\begin{figure}

\begin{center}
\newcommand{\ige}[1]{\includegraphics[height=2.5cm]{figs/baseline_match_examples/#1.jpeg}}
\newcommand{\imcredsF}[1]{\raisebox{\depth}{\scalebox{0.4}{\rotatebox{270}{Flickr / #1}}}}
\begin{tabular}{cc@{}c}
\multicolumn{3}{c}{GIST PCA 256}\\
\ige{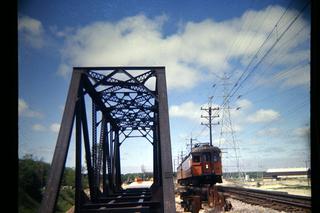} & \ige{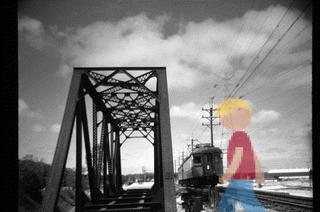} & \imcredsF{davidwilson1949}\\
\ige{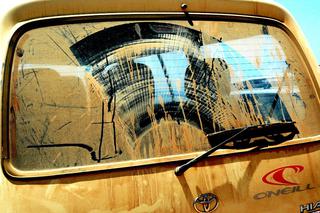} & \ige{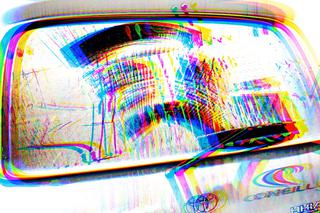} & \imcredsF{Gustty}\\
\multicolumn{3}{c}{Multigrain + score normalization}\\
\ige{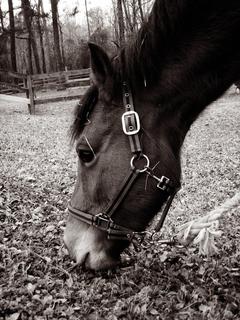} & \ige{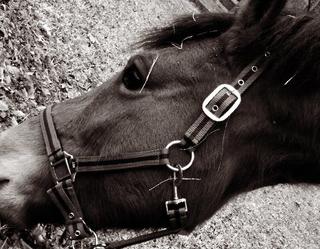} & \imcredsF{Miss Skew}\\
\ige{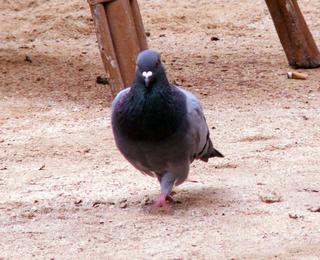} & \ige{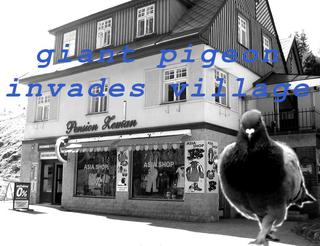} & \imcredsF{YuvalH}\\
\ige{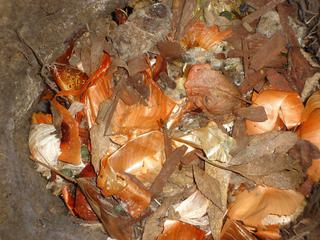} & \ige{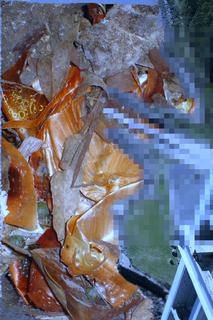} & \imcredsF{mlinksva}\\
\multicolumn{3}{c}{HOW + score normalization}\\
\ige{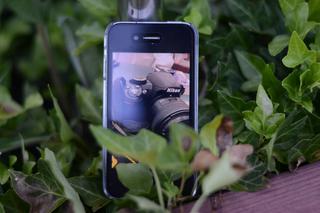} & \ige{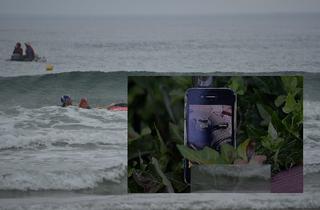} & \imcredsF{araza123}\\
\ige{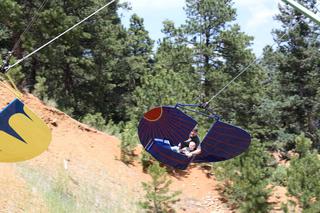} & \ige{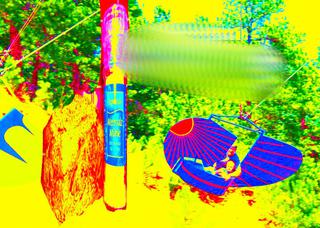} & \imcredsF{carlock\_family}\\
\ige{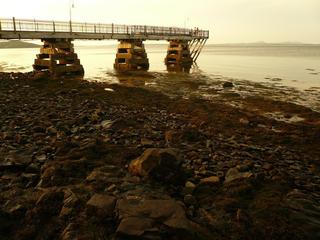} & \ige{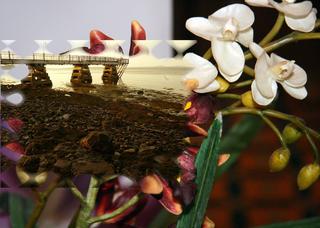} & \imcredsF{haiden1991}\\

\end{tabular}
\end{center}
    \caption{
    Examples of difficult image pairs that each of the baseline methods are able to match in a regime of above 90\% precision. 
    Left: source image, right: image match.
    }
    \label{fig:examplematches}
\end{figure}

Figure~\ref{fig:examplematches} shows typical examples where each of the baseline descriptors produces reliable matches ($>$ 90\% precision). 
GIST can handle light color changes and small crops. 
HOW+ASMK is particularly good at matching strong crops and cluttered scenes, a domain classically dominated by local descriptor methods. 
False positive results (not showed here) suggest that especially HOW and Multigrain are too invariant to aspect changes like different viewpoints of 3D objects.

\section{The competition}
\label{sec:rules}

The competition is intended to be as open as possible to individual, academic and industrial participants.

The official submission site is managed by Driven Data, via the website~\url{https://www.drivendata.org/competitions/79/}. 
It provides the dataset, the evaluation script with partial ground truth, the leaderboard and the submission site. 

The competition rules and precise file formats are described on the website, this section is intended as a readable digest.

\subsection{Participants}

A participant is a group of people that can submit results to the competition site. 
We disallow (1) people to move from one group to another and (2) private communication between groups -- any information provided by one group should be made available to all other groups as well. 

\paragraph{Open vs. closed-source}

Participants can choose to register either as open-source or as closed-source. 
For the open-source version, participants promise to release the source code prior to the final submission.
For the closed-source version, participants do not need to submit source code, but they will not be considered for the prizes.

\subsection{Tracks}

\label{sec:tracks}

There are two tracks in the competition, participants can choose to compete in either or both of the tracks.

\paragraph{Track 1.}

Participants submit lists of matching images as a CSV file. 
The image matches are reported as a triplet of (query image, reference image, score). 
The source and reference images are the image file names. 
The score is a floating point number that evaluates how confident the match is (higher is more confident).
The submission is evaluated by computing the micro-AP measure on the result list (see Section~\ref{sec:metric}). 

\paragraph{Track 2.}

Participants submit global image descriptor vectors for the query images and the reference images as a HDF5 file.
The image descriptors are in 256 dimensions at most.
The matches are performed using L2 distance on the evaluation server. 
Then the submission is evaluated in the same way as for Track 1. 
This setting is more constrained, therefore we expect the track 2 results to always be lower or equal to the track 1 results. 

\subsection{Phases}

The competition will take place in three phases.

\paragraph{Phase I: development.}

(between June and October).
The participants can register, tinker with the dataset, use the provided ground truth, and submit to the development phase leaderboard. 

The organizers provide the reference set, and the development query set. Ground-truth correspondences of the queries with the reference images are provided only for the 25,000 of the query images.
For the other 25,000 images, the ground-truth is not provided but they are taken into account for the evaluation server.

The ``open-source'' participants should submit the code of their method prior to the evaluation phase. 
In addition, all Track 2 participants should submit the final descriptors for the reference images.

\paragraph{Phase II: evaluation.}

(48h in October). 
The test query set is delivered to the users, without the ground-truth.
Users are allowed 3 submissions of search results on the final queries.

\paragraph{Phase III: verification.}

(November) 
The organizers verify that the code submitted in advance reproduces the reported results. 
The participants are expected to reply to requests from the organizers to assist in this validation.

The results will be announced at the NeurIPS competition workshop.

\subsection{Provided material}

The organizers provide the following code to participants. 

\paragraph{Evaluation script.}

This script is the one used for the track~1 and track~2 evaluation, computing the global AP metric. 
It can be run locally by participants on the partial ground truth on 25,000 query images. 

\paragraph{Baseline methods.}

The code for baseline methods (Section~\ref{sec:baselines}) is provided. 
It is intended for validation and as a starting point for participants, on which they can build to develop their methods. 

\subsection{Acceptable methods}

The scoring of a query image w.r.t. a reference image should be independent of other query images and reference images. 
This means that even if the dataset had a single query image and a single reference image, the score of the image pair would be the same. 
The intent of this rule is to avoid 
(1) that algorithms overfit to the reference set, for example by building a gigantic classifier with 1M outputs that predicts the matches, 
(2) that algorithms use irrelevant dataset statistics like the fact that there is at most one query image per reference image. 
This rules out methods based on query expansion~\cite{chum2007total} or neighborhood graphs~\cite{qin2011hello}.

The training set is provided as a statistical twin of the reference set, it can be used to do all kinds of training tasks without risk of overfitting to the reference set. 

We ask participants to provide some meta-information on their runs, like the processing time and memory usage. 
This is just for information purposes, higher resource consumption is not penalized.

\section{Conclusion}
\label{sec:conclusions}

The \OURD dataset aims at probing the state of the art of image de-duplication. 
We paid particular attention to the size of the dataset and calibration of the difficulty of transformations. 
We also did our best to ascertain that there are no loopholes that participants could take advantage of. 
We hope that many participants will take part in the challenge and that this will re-fuel the interest in the near-duplicate detection task. 
In the long run, we will make sure that the data will remain available on the website \url{https://github.com/facebookresearch/isc2021} so that progress can be measured over a few years.

\bibliographystyle{abbrv}
\bibliography{biblio}

\newpage
\appendix

\section{Additional transformations for the final phase}
\label{sec:moretransfo}

After the development phase of the competition, it appeared that the top submissions were stronger than we anticipated ($\mu AP>$90\% in the matching track). 
Therefore we decided to make the final phase transformations harder than those of the development phase. 
In addition, as decided beforehand, we included a fraction of new transformations, to simulate the realistic setting where the attacks are not completely known at training time and avoid overfitting to the development phase attacks.

\subsection{Automatic transformations}

The automatic transformations are composed of a sequence of elementary AugLy transformations~\cite{augly} sampled randomly. 
We increased the likelihood of applying more transformations sequentially: whereas in the development set we applied 1-4 transformations to create each query image with increasing probabilities for higher numbers, for the test set we applied 1-5 and skewed the probabilities even more heavily so that it was much more likely to apply 3-5 transformations than 1-2.

We added some new transformations from AugLy to the set we applied:
\begin{itemize}
      \item Overlays: overlaying stripes onto the image
      \item Pixel-level: adding random noise, ``legofy'' (a type of pixelization which replaces a number of adjacent pixels with a lego block of the average color)
      \item Spatial: vertical flip, adding other images in a collage around the image
\end{itemize}

We sampled harder parameters for some transformations compared to the dev set including blur, encoding\_quality and pixelization. 
For example, for blur the radius is sampled in $[5, 10]$ for the development phase and in $[5, 15]$ in the final phase.

\subsection{Air-gap}

\begin{figure}[t]
\begin{center}
\newcommand{\ige}[1]{\includegraphics[height=2.5cm]{figs/disc_images/#1}}
\newcommand{\ecom}[1]{#1}
\newcommand{\imcreds}[1]{\raisebox{\depth}{\scalebox{0.4}{\rotatebox{270}{Flickr / #1}}}}
\newcommand{\imcredsD}[1]{\raisebox{\depth}{\scalebox{0.4}{\rotatebox{270}{DFDC image #1}}}}
\begin{tabular}{cc@{}c}
\multicolumn{3}{c}{\ecom{Air-gap transform}}\\
\ige{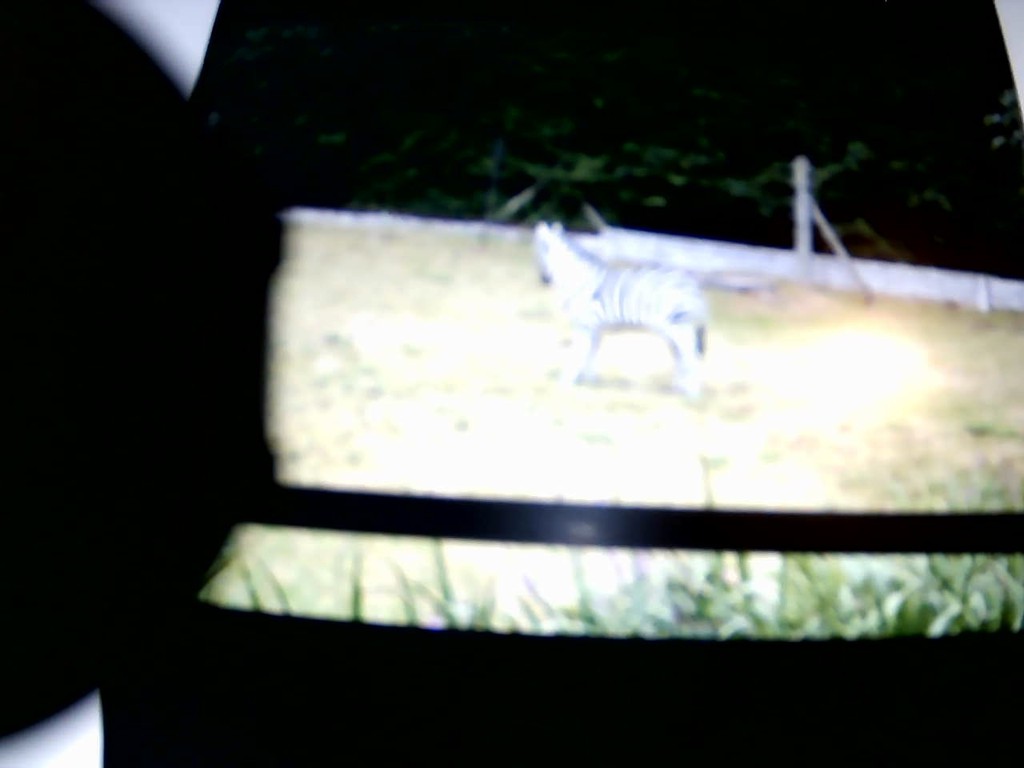} & \ige{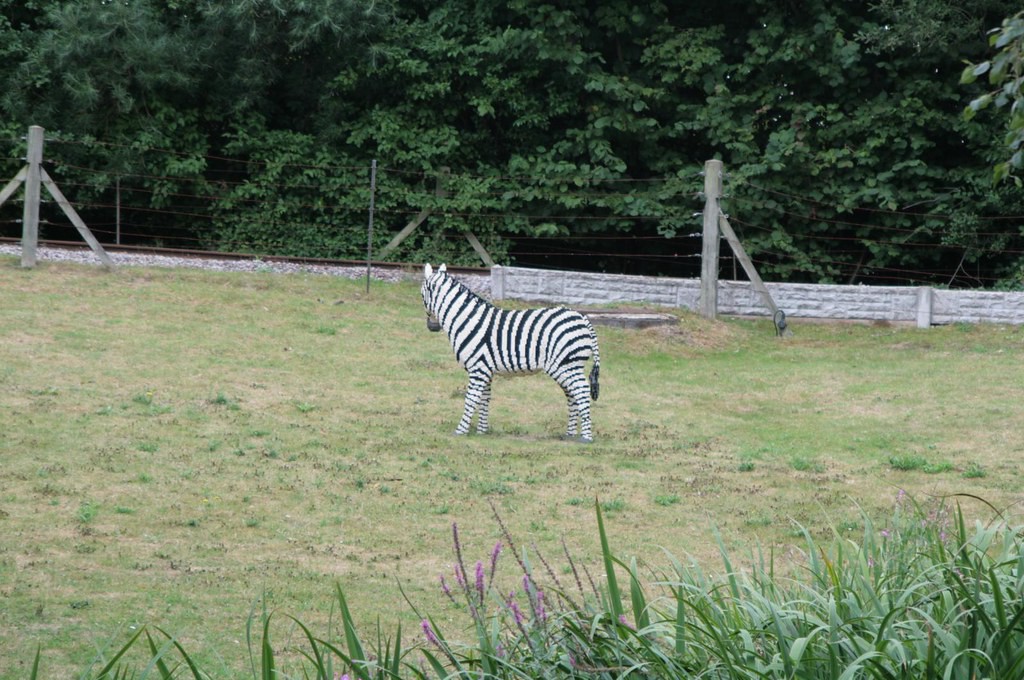} & \imcreds{Karen Roe}\\
\multicolumn{3}{c}{\ecom{Air-gap transform}}\\
\ige{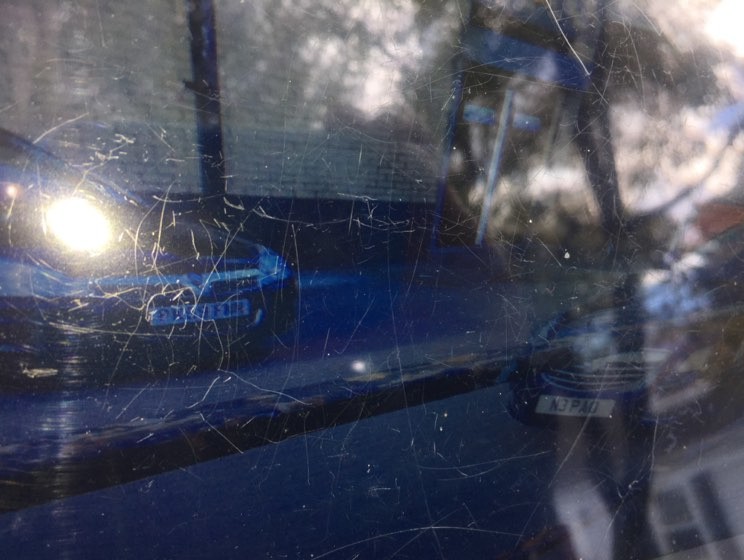} & \ige{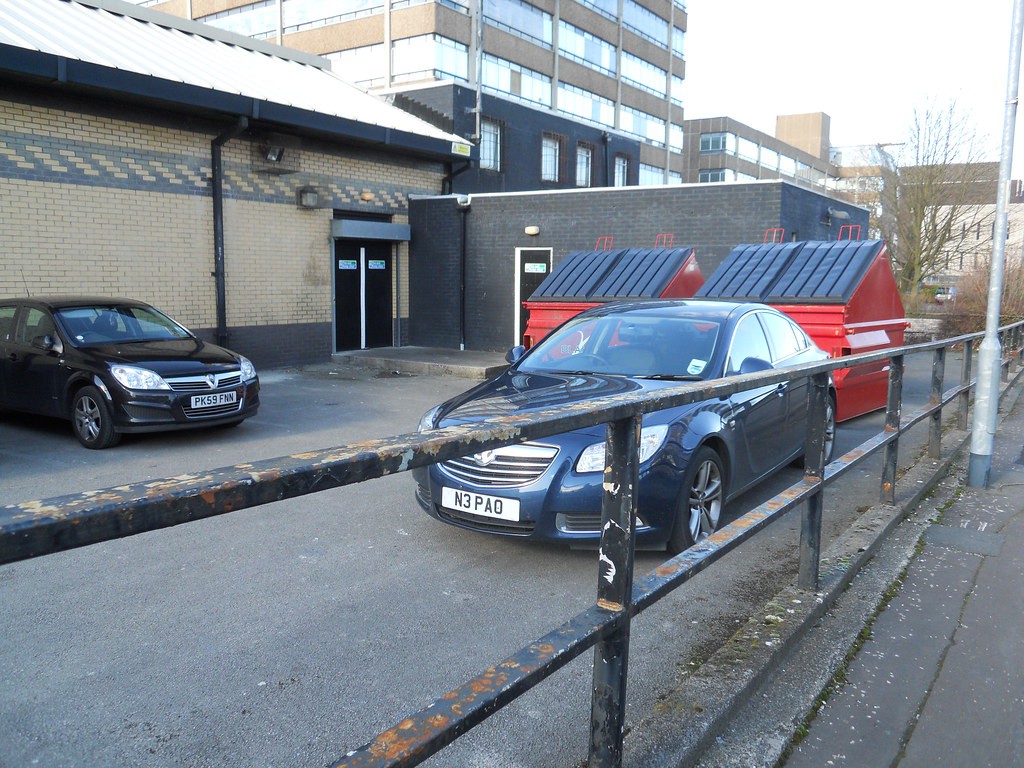} & \imcreds{Firing up the quattro....}\\
\multicolumn{3}{c}{\ecom{Legofy}}\\
\ige{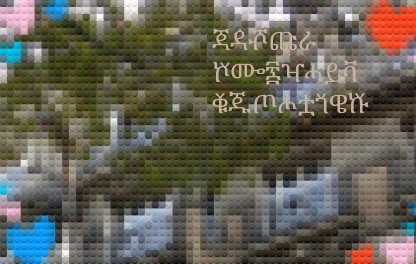} & \ige{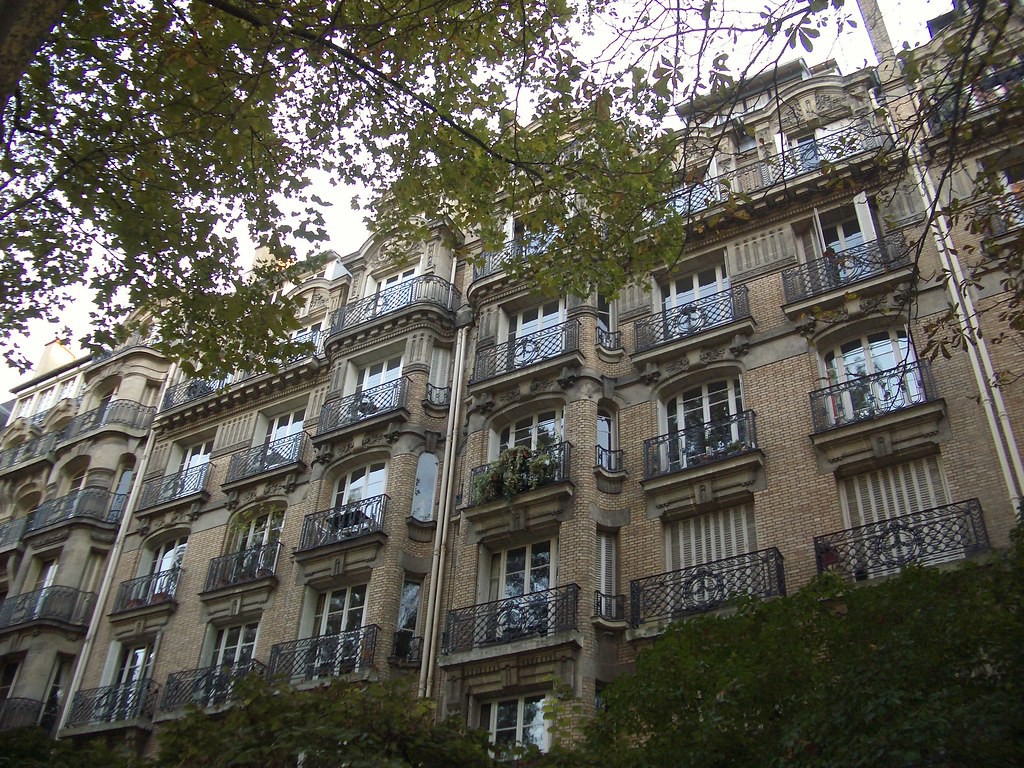} & \imcreds{lamouridanielle}\\
\multicolumn{3}{c}{\ecom{stripes + adversarial attack, resnet34, PSNR=20}}\\
\ige{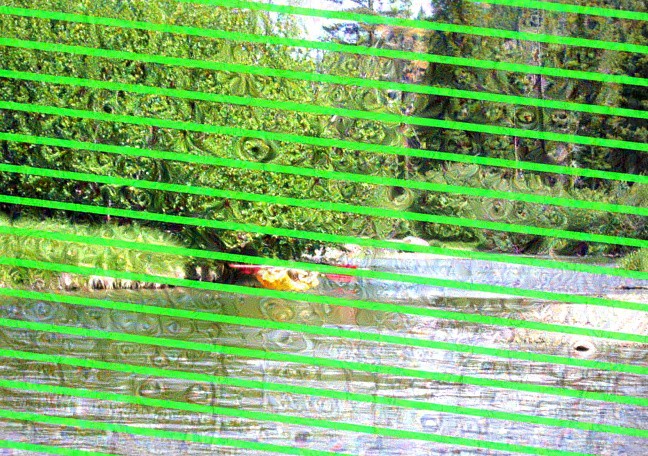} & \ige{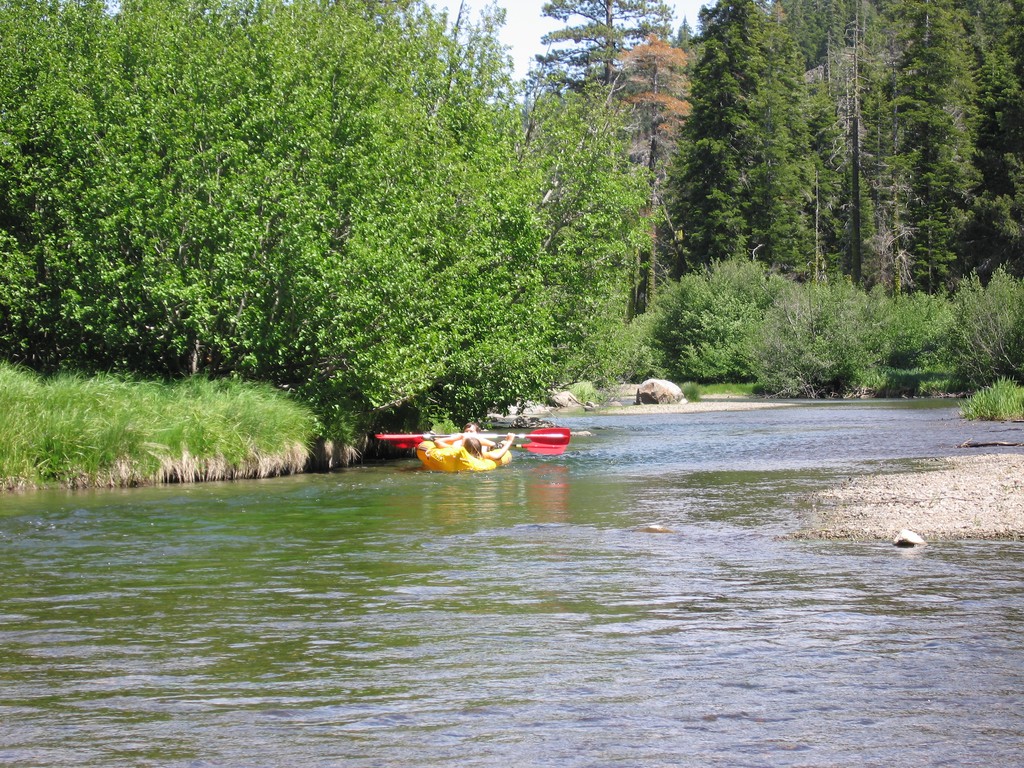} & \imcreds{Vlad \& Marina Butsky}\\
\multicolumn{3}{c}{\ecom{Adversarial attack, vgg19, PSNR=20}}\\
\ige{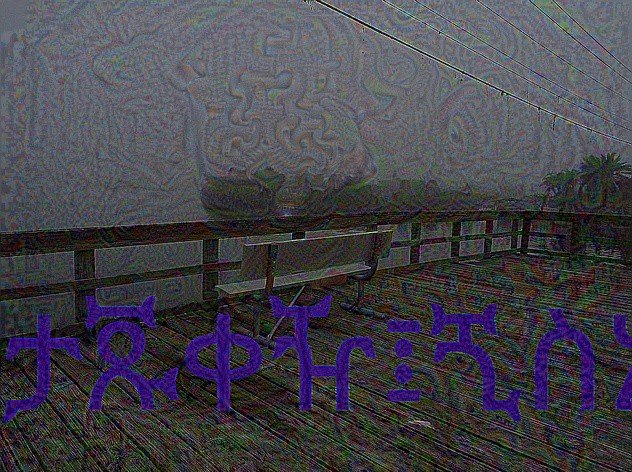} & \ige{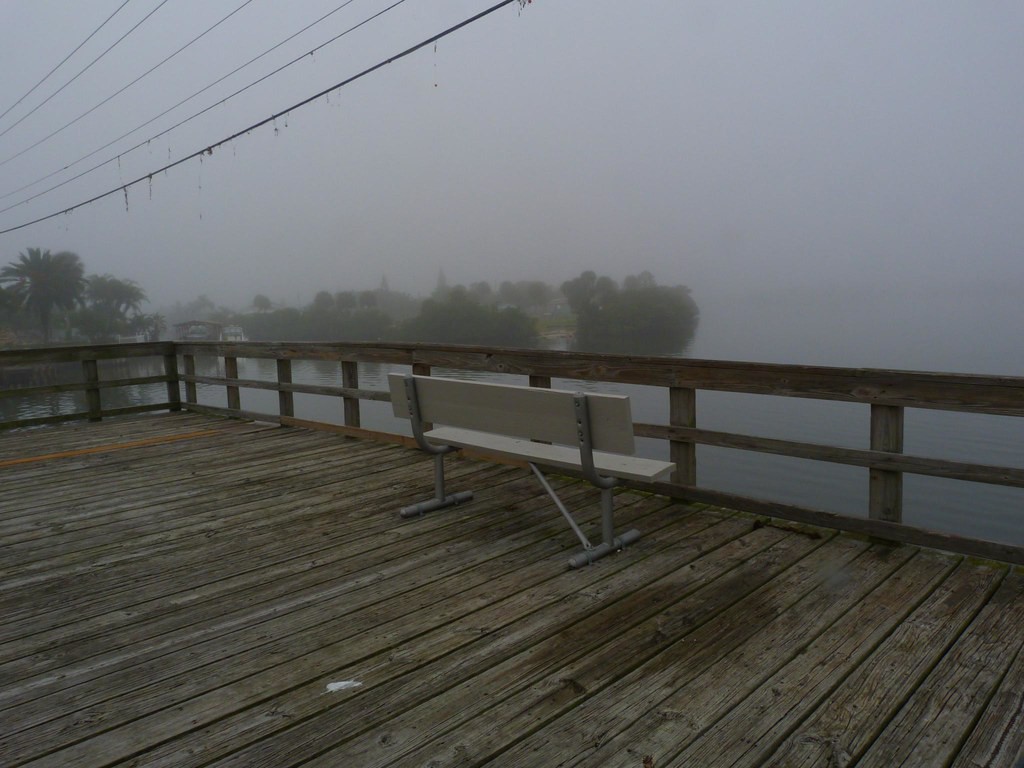} & \imcreds{Rusty Clark}\\
\end{tabular}
\end{center}
\caption{\label{fig:additionaltransfo}
Example transformations for the final phase of the competition.
}
\end{figure}

We asked manual editors to perform ``air-gap'' transformations, where the picture is displayed or printed out, and re-captured with a camera or mobile phone. 
It is a quite common image attack, that is hard to reproduce automatically or even with an image editor like the GIMP.
Figure~\ref{fig:additionaltransfo} shows an example air-gap augmentation (the zebra). 
The effect is a combination of over-exposure, geometric framing and blurriness on the picture.

\subsection{Adversarial transformations}

\begin{figure*}
    \centering
    \includegraphics[width=\linewidth]{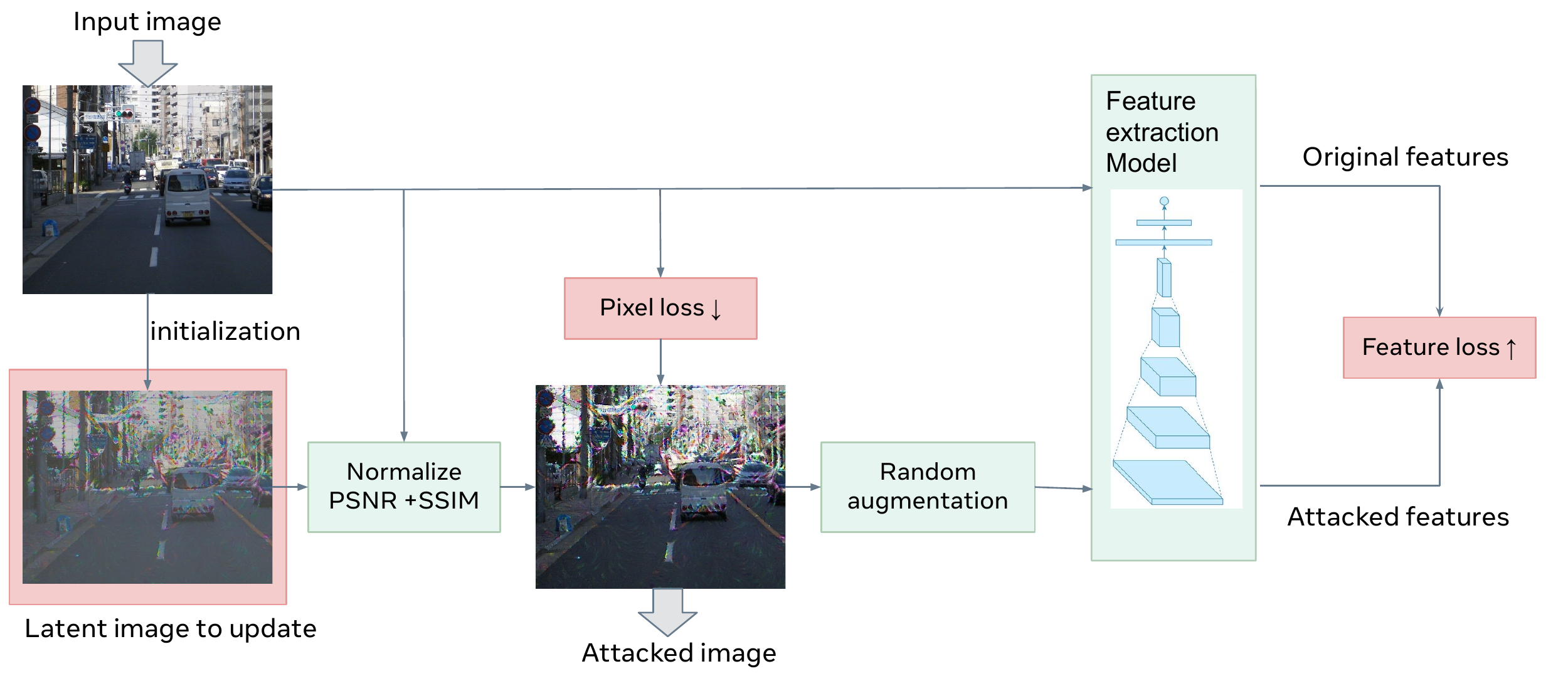}
    \caption{
        Adversarial attack optimization.
    }
    \label{fig:adv_attacks}
\end{figure*}

Since is is likely that most submissions to the challenge are based on an image analysis with neural nets, we developed an attack that specifically targets those.
This is inspired by previous works that show that even imperceptible changes to images can change the classification accuracy~\cite{szegedy2013intriguing,goodfellow2014explaining,shamir2019simple} or a retrieval system~\cite{tolias2019targeted}.
Some of these methods can even be applied with only black-box access to the models~\cite{papernot2017practical}. 
However, this does not help here because we don't have access to the participant's models, even in black box. 
Instead, we start from a hypothesis of what model $f$ the participants \emph{could be using} and attack that model in white-box mode. The goal is to slightly perturb an image so that when it is transformed its features will not be matching those of the original image.

\paragraph{Optimization.}

Our approach is based on the techniques used in watermarking~\cite{fernandez2021watermarking} because they offer a fine control of the distortion on the image. 
The attack starts from the source image $x$ and produces an image $y$ by distorting $x$. 
It minimizes the following loss that combines a pixel difference term (that must be minimized) and a feature difference term (that should be maximized):
\begin{equation}
    \mathcal{L} = 
    w \| x - y \|^2
    - 
    (1 - w) \| f(x) - f(T(y)) \|^2, 
\end{equation}
where $w\in[0,1]$ is a weight factor and $T$  is a random transformation applied to the image (we chose $w=1/2$ for all experiments). 
A different transformation $T$ is sampled at each iteration of the optimization. 
It is intended to make the optimization less sensitive to other image transformations that could be applied. 
The optimization does not operate on the model parameters or on $y$ directly but on the pixels of a latent image $z$ that is subsequently normalized 
\begin{equation}
    y = N(\mathrm{SSIM}(z, x), x, p)
\end{equation}
The first normalization step $\mathrm{SSIM}$~\cite{wang2004image} makes sure the changes w.r.t. the original image $x$ occur preferably on edges (high frequency) of the image content. 
The second normalization step $N$ scales the difference image between $z$ and $x$ so that the PSNR between the two remains above a target PSNR $p$: 
\begin{equation}
    N(z, x, p) = x + \frac{\varepsilon(p)}{\| z - x \|} (z - x) 
\end{equation}
where $\varepsilon(p)$ is a decreasing function of $p$. 

\begin{equation}
    \varepsilon(p) = 10^\frac{K-p}{20} 
\end{equation}
where $K = 10\log_{10}(255) + 10 \log_{10} N_\mathrm{pix}$, $N_\mathrm{pix}$ being the number of pixels in the image. 

We use 100 iterations of the Adam optimizer with a learning rate of 0.05 to propagate back to the latent image pixels $z$ to minimize the loss $\mathcal{L}$. 
Figure~\ref{fig:additionaltransfo} shows a few examples of attacked images with target PSNRs of 20. 
A transformation with PSNR=20 is very visible, while 40 is perceivable for trained eyes.

\paragraph{Impact of adversarial attacks.}

We evaluate the adversarial attacks on the baseline methods of the challenge. 
We use several levels of PSNR and pretrained models. 

We measure the $\mu AP$ on the subset of development queries that have a match. 
This is the simplest setting of the development kit of the challenge, without any score normalization. 
The SSCD model~\cite{pizzi2022sscd} is a resnet50 trained with self-supervised learning (similar to \cite{chen2020simple}) on the DISC2021 training set.

Table~\ref{tab:aa_baselines} shows the results. 
Multigrain is very sensitive to attacks from all models. 
It is based on a resnet50 model trained on imagenet so it is not surprising that the resnet50-based attack is the most efficient. 
The local descriptor method HOW~\cite{tolias2020learning} is much less sensitive to adversarial attacks, despite being based on the same resnet50 model. 
SSCD is trained with a specific entropy loss term that probably makes it less sensitive to attacks, especially from other models. 
Even the SSCD based attack (the only white box attack in these experiments) has an impact of less than 0.1 on the AP. 
It is interesting to notice that the behavior of the resnet50 attack is closer to Dino~\cite{caron2021emerging} than to VGG16, despite the fact that Dino is trained in a self-supervised way on YFCC100M instead of Imagenet.
The SSCD attack is less efficient on the other approaches \ie it is not a good attack model.

\begin{table}[]
    \centering
    \begin{tabular}{lr|rrr}
             &       & \multicolumn{3}{c}{Retrieval method} \\
Attack model & PSNR      & Multigrain  & HOW         & SSCD  \\
\hline
None         & $\infty$ 
                  & 0.563       & 0.500       & 0.812 \\
resnet50,    & 30 & 0.120 & 0.270  & 0.779 \\
resnet50,    & 40 & 0.329 & 0.431  & 0.802 \\
Dino (deit-S)& 40 & 0.332 & 0.425  & 0.794 \\
VGG16        & 40 & 0.363 & 0.446  & 0.804 \\ 
SSCD         & 40 & 0.447 & 0.466  & 0.719 \\
    \end{tabular}
    \caption{Impact on adversarial attacks targeting different feature extraction models on the baseline methods of the challenge.
    The performance is measured as $\mu AP$ on a subset of the development queries.
    }
    \label{tab:aa_baselines}
\end{table}

\end{document}